\documentclass[10pt,twocolumn,letterpaper]{article}

\usepackage{cvpr}              
\definecolor{cvprblue}{rgb}{0.21,0.49,0.74}
\usepackage[pagebackref,breaklinks,colorlinks,allcolors=cvprblue]{hyperref}

\usepackage{hyperref}
\usepackage{url}
\usepackage{kotex}
\usepackage{booktabs}
\usepackage{multirow}
\usepackage{makecell}
\usepackage{graphicx}
\usepackage{amsmath}
\usepackage{amssymb}
\usepackage{algpseudocode}
\usepackage{capt-of}
\usepackage{subcaption}
\usepackage{graphicx}
\usepackage{booktabs}
\usepackage{capt-of}
\usepackage{adjustbox}
\usepackage[table]{xcolor}

\definecolor{ablBlue}{HTML}{2465E8}
\definecolor{ablPurple}{HTML}{923BE5}
\definecolor{ablOrange}{HTML}{D78B00}
\definecolor{ablRed}{HTML}{EF3340}
\definecolor{ablTeal}{HTML}{009D87}

\usepackage{times}
\usepackage{epsfig}
\usepackage{float}
\usepackage{placeins}
\usepackage{enumitem}
\usepackage{tabularx}
\usepackage{xstring}
\usepackage{xspace}
\usepackage[hang,flushmargin]{footmisc}
\usepackage[utf8]{inputenc} 
\usepackage[T1]{fontenc}    
\usepackage{url}            
\usepackage{booktabs}       
\usepackage{amsfonts}       
\usepackage{nicefrac}       
\usepackage{xcolor}         
\usepackage{amsmath, amssymb, overpic, textpos}
\usepackage{graphicx}
\usepackage{multirow}
\usepackage{color, colortbl}
\usepackage{tabulary}
\usepackage{listings}
\usepackage{multicol}
\usepackage{xspace}
\usepackage{graphbox}
\usepackage{arydshln}
\usepackage{adjustbox}
\usepackage{kotex}
\usepackage{caption}
\usepackage{subcaption}
\usepackage{makecell}
\usepackage{tabularray}
\usepackage{bm}
\usepackage{wrapfig}

\definecolor{Green}{rgb}{0.2, 0.7, 0.1}
\definecolor{Gray}{HTML}{B0B0B0} 
\definecolor{Blue}{HTML}{4A90E2} 
\definecolor{Yellow}{HTML}{FFA700} 
\definecolor{prompt_blue}{HTML}{1f78b4}
\definecolor{prompt_red}{HTML}{d45c43}
\definecolor{plus}{HTML}{0071bc}
\definecolor{minus}{RGB}{153,10,10}
\definecolor{SecondBest}{HTML}{E0F0FA}
\definecolor{Best}{HTML}{BAD8F2}

\usepackage[table]{xcolor}

\usepackage[utf8]{inputenc}

\usepackage{tcolorbox}
\usepackage{etoc}
\usepackage{titletoc}
\definecolor{blush}{RGB}{220, 70, 130}
\usepackage{multirow}
\usepackage[table]{xcolor}
\usepackage{array}  

\usepackage[ruled,vlined]{algorithm2e}
\definecolor{cvprblue}{rgb}{0.21,0.49,0.74}
\usepackage[accsupp]{axessibility}
\usepackage{xcolor}

\definecolor{blush}{RGB}{220, 70, 130}

\newcommand\bestrob[1]{\textbf{\color{blush}#1}}
\newcommand\bestclean[1]{\textbf{#1}}
\usepackage{booktabs}
\definecolor{lightergray}{gray}{0.94}
\usepackage{booktabs}
\usepackage[table]{xcolor}
\usepackage{pifont}
\usepackage{amsmath}

\definecolor{lightrow}{RGB}{245,239,210}

\newcommand{\mc}[2]{\multicolumn{#1}{c}{#2}}

\newcolumntype{a}{>{\columncolor{lightergray}}c}

\def\confName{CVPR}
\def\confYear{2027}

\title{T-VSS: Test-Time Visual Subspace Steering for Adversarial Robustness of Vision-Language Models}

\author{%
    Jaehyuk Jang \quad
    Minseok Seo \quad
    Seungju Cho \quad
    Kangwook Ko \quad
    Changick Kim \\[0.3em]
    Korea Advanced Institute of Science and Technology, South Korea \\
    \texttt{\{jhyuk, minseok.seo, joyga, kw.ko, changick\}@kaist.ac.kr} \quad
}

\begin{document}
\maketitle

\begin{abstract}

Contrastive vision-language models (VLMs), exemplified by CLIP, enable strong zero-shot visual recognition but remain highly vulnerable to adversarial perturbations.
Recent test-time adaptation methods improve robustness without retraining.
Prompt-based methods adapt learnable text prompts, while input-space methods optimize pixels or padding at test time.
These approaches can improve predictions, but they do so through an indirect and expensive optimization path.
We propose \textbf{Test-time Visual Subspace Steering (T-VSS)}, a lightweight defense that performs test-time adaptation directly in the visual feature space.
T-VSS first builds a sample-specific low-rank subspace from multi-view feature residuals anchored at the attacked image.
It then learns a shared feature correction within this subspace using reliability-weighted entropy minimization.
By constraining adaptation to a compact visual geometry, T-VSS steers attacked features toward more stable and discriminative predictions while avoiding noisy full-space updates.
Experiments on fine-grained, ImageNet, and ImageNet-OOD benchmarks show that T-VSS improves adversarial robustness while maintaining competitive clean accuracy and better efficiency than prior test-time adaptation methods.
Code is available at \href{https://github.com/jhyukjang/T-VSS}{here}.

\end{abstract}


\section{Introduction}

Vision-language models (VLMs)~\cite{radford2021learning,chen2023vlp,zhang2024vision,sun2023eva} have become a strong foundation for zero-shot visual recognition.
By aligning images and text in a shared embedding space, they can recognize unseen categories and transfer to downstream tasks without task-specific fine-tuning.
This flexibility makes VLMs attractive in realistic settings where labeled data are scarce and rapid deployment is important.

However, their zero-shot predictions are highly sensitive to adversarial perturbations, where even small and nearly imperceptible input changes can lead to incorrect predictions~\cite{goodfellow2015explaining,madry2018towards,fang2024clip,li2024language,schlarmann2024robust}.
This vulnerability is particularly concerning in safety-critical applications such as medical AI, autonomous driving, and public security surveillance, where small perceptual failures can lead to high-stakes downstream consequences~\cite{finlayson2019adversarial,eykholt2018robust,ijcai2021p591}.
Classical approaches such as adversarial training~\cite{madry2018towards,zhang2019theoretically,cui2026agft}, robust fine-tuning~\cite{mao2023understanding,schlarmann2024robust,wang2024pre,zhang2026semanticaware}, diffusion-based purification~\cite{nie2022DiffPure}, and adversarial prompt tuning~\cite{li2024one,zhang2024adversarial,zhou2024few} can improve robustness, but they typically require additional data, label access, expensive retraining, or auxiliary models.
To avoid these costs, recent work has explored test-time adaptation, which seeks to improve robustness using only unlabeled inputs at inference.

\begin{figure*}[t!]
    \centering
    \includegraphics[width=.85\linewidth]{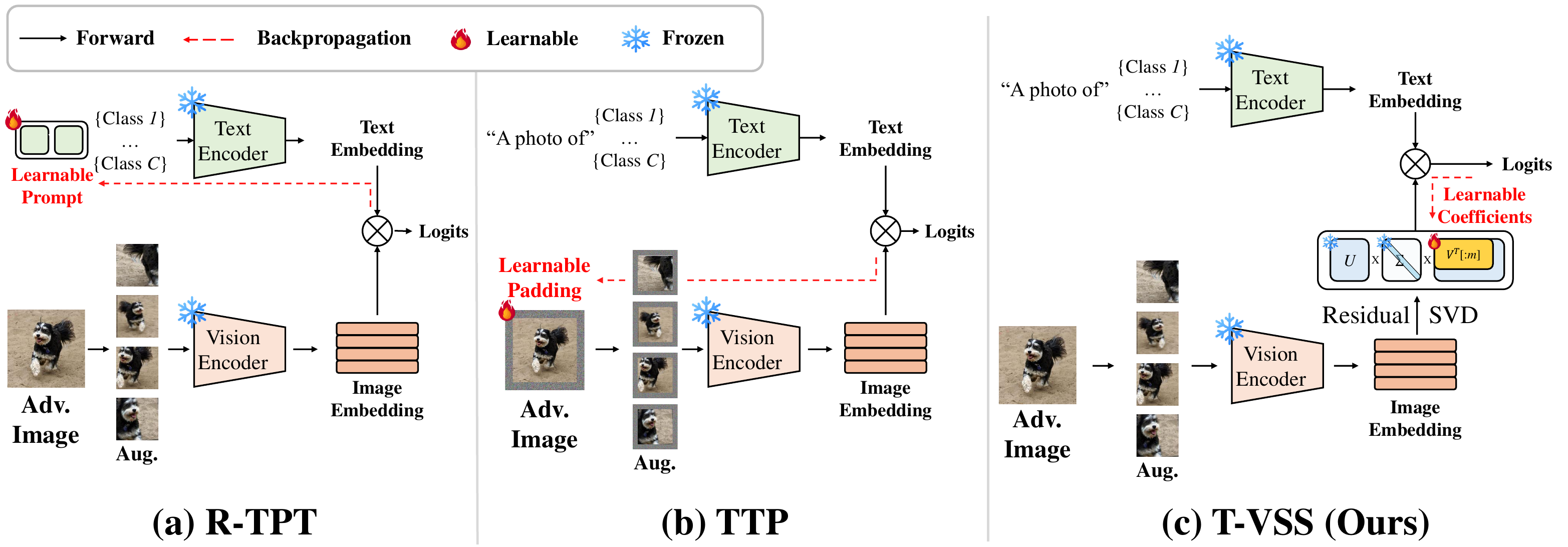}
    \vspace{-0.15cm}
    \caption{
Comparison of test-time adaptation strategies for adversarially robust vision-language models.
(a) \textbf{R-TPT}~\cite{sheng2025r} adapts learnable text prompts through backpropagation in the text branch.
(b) \textbf{TTP}~\cite{ttp} optimizes learnable input padding in the pixel space via backpropagation through the vision encoder.
(c) In contrast, \textbf{T-VSS} adapts the visual feature space by learning a small set of steering coefficients in a low-rank visual subspace.
Compared with prior methods, T-VSS provides a more direct and lightweight correction mechanism.
}
    \vspace{-0.15cm}
    \label{fig:comparison}
\end{figure*}

In the VLM setting, representative methods adapt each test sample either through the text branch or in the input space at test time as illustrated in Fig.~\ref{fig:comparison}: R-TPT~\cite{sheng2025r} improves adversarial robustness by optimizing learnable prompts, whereas TTC~\cite{xing2025clip} and TTP~\cite{ttp} operate in the input space by optimizing a learnable counterattack perturbation and instance-specific image padding, respectively.
Despite their promise, these methods improve predictions only through indirect and often computationally expensive optimization.
This indirection is problematic because adversarial attacks ultimately impair zero-shot recognition by corrupting the visual feature that is matched against text prototypes.
A more direct defense should therefore adapt the attacked visual representation itself.
However, unconstrained feature-space adaptation can be unstable.
If the update is not guided by the local structure of the test sample, it may move the feature away from its semantic content and reinforce an incorrect prediction.
The key challenge is therefore to steer the attacked representation at test time toward a more discriminative region while restricting the update to 
sample-specific directions derived from augmented views.

In this paper, we address this challenge with \textbf{Test-time Visual Subspace Steering (T-VSS)}, a lightweight defense that adapts each sample within a compact, sample-specific visual subspace.
Given a test image and its augmented views, the method extracts frozen visual features once and forms anchor-based residuals by subtracting the original test-image feature from each augmented-view feature.
Our design is motivated by two complementary properties of attacked multi-view features.
First, perturbation-induced feature shifts remain partially aligned across views, suggesting that a shared correction can benefit multiple transformations of the same attacked image.
Second, feature residuals measured relative to the original
attacked view concentrate most of their energy in a few
directions. These observations play distinct roles:
cross-view alignment motivates a shared correction,
whereas residual energy concentration motivates a compact,
sample-specific search space.
Accordingly, T-VSS applies singular value decomposition to the anchor-relative residuals and optimizes a shared correction only within the resulting subspace.
Because augmented views can vary in reliability under attack, it estimates view reliability from feature-level agreement across views.
The resulting weights are used during both adaptation and final aggregation, reducing the influence of unstable views on the final prediction.
By constraining entropy minimization to these dominant residual directions, T-VSS reframes adversarial test-time adaptation as structured feature correction rather than prompt tuning or dense pixel-space search.
This change in adaptation space has a direct empirical payoff: the proposed approach consistently improves the robustness--efficiency trade-off over prior test-time defenses.

Extensive experiments across eight fine-grained datasets, ImageNet, and four ImageNet-OOD benchmarks using multiple CLIP~\cite{radford2021learning} backbones demonstrate superior average adversarial robustness while maintaining competitive clean accuracy.
Since optimization is performed only over a small set of subspace coefficients rather than text prompts or dense input variables, the method also reduces inference overhead.
These results position compact feature-space steering as a simple and practical alternative to prompt- and pixel-space adaptation for robust zero-shot VLM inference.

\section{Related Work}

\subsection{Adversarial Attacks and Defenses}

Adversarial examples reveal the vulnerability of deep neural networks by introducing small, often imperceptible perturbations that lead to incorrect predictions~\cite{goodfellow2015explaining, kurakin2018adversarial, madry2018towards, carlini2017evaluating}. 
Representative attacks include single-step methods such as FGSM~\cite{goodfellow2015explaining}, iterative optimization-based methods such as BIM and PGD~\cite{kurakin2018adversarial, madry2018towards}, and stronger transfer-based or black-box attacks that do not require direct gradient access.
Universal perturbations further demonstrate that a shared perturbation can fool a model across many inputs~\cite{moosavi2017universal}. 
Recent studies show that vision-language models (VLMs), despite their strong zero-shot generalization, are also highly vulnerable to such attacks, which poses a major obstacle to reliable deployment.

To mitigate this issue, prior defenses have mainly focused on training-time robustness improvement. 
Adversarial training and its variants improve robustness by explicitly optimizing models on adversarially perturbed examples~\cite{madry2018towards, zhang2019theoretically, rice2020overfitting,cui2026agft}, while robust fine-tuning~\cite{mao2023understanding,schlarmann2024robust, wang2024pre,zhang2026semanticaware}, diffusion-based purification~\cite{nie2022DiffPure}, and adversarial prompt tuning~\cite{li2024one,zhang2024adversarial,zhou2024few} extend this paradigm to VLMs. 
Although effective, these methods typically require labeled data, repeated adversarial example generation, and costly retraining or fine-tuning of large pretrained models. 
Such requirements are especially burdensome for large frozen VLMs, motivating lightweight defense mechanisms that can operate directly at inference time.

\subsection{Test-Time Adaptation and Defense for VLMs}

Test-time adaptation (TTA)~\cite{nado2021evaluatingpredictiontimebatchnormalization,seo2026efficient,liu2021ttt,sun19ttt} aims to improve model generalization on unseen test distributions using only unlabeled test inputs. 
For VLMs, Test-Time Prompt Tuning (TPT)~\cite{shu2022test} tunes learnable text prompts~\cite{zhou2022conditional, zhou2022learning} using multiple augmented views of each test image, and follow-up methods such as MTA~\cite{zanella2024test} further improve adaptation stability through multi-view or multi-prompt aggregation.
STS~\cite{sts} instead performs spectrum-aware latent steering in the text embedding, enabling efficient adaptation. However, these methods are primarily designed for natural distribution shifts and generally assume clean test inputs, which limits their effectiveness under adversarial perturbations.

Recent work has extended TTA to adversarially robust inference for VLMs. 
R-TPT~\cite{sheng2025r} revisits TPT under adversarial attack by optimizing learnable text prompts with pointwise entropy over selected low-entropy views at test time.
A different line of work instead adapts the input itself at test time.
%
TTC~\cite{xing2025clip} and ET3~\cite{ET3} perform test-time defenses by optimizing an additive perturbation in the image space, while TTP~\cite{ttp} optimizes instance-specific padding parameters.
Separately, recent training-free defenses bypass optimization-based adaptation, leveraging textual descriptions generated by large language models~\cite{cola}, or applying calibration-dependent thresholded feature reconstruction~\cite{liu2026adversarial}.

In this paper, we focus on optimization-based test-time adaptation.
Our T-VSS operates directly on attacked visual representations by learning a shared correction in a sample-specific low-rank visual subspace, providing a geometry-aware alternative to both prompt-space adaptation and padding-based input optimization without auxiliary models or datasets.

\section{Method}
\label{sec:method}

\subsection{Preliminaries}

\paragraph{CLIP for Zero-Shot Classification.}
We build on CLIP~\cite{radford2021learning}, a dual-encoder vision-language model consisting of an image encoder $F(\cdot)$ and a text encoder $G(\cdot)$. 
Given a $C$-way classification task with class names $\{t_c\}_{c=1}^C$, CLIP constructs a text prototype for each class as $g_c = G(\mathrm{prompt}(t_c)) \in \mathbb{R}^d$, where $\mathrm{prompt}(t_c)$ denotes a hand-crafted prompt template (e.g., ``a photo of a [CLASS]'') instantiated with class name $t_c$, and $g_c$ is the resulting textual embedding of class $c$.
For an input image $x_i$, the image encoder produces a visual feature $f_i = F(x_i) \in \mathbb{R}^d$.
The zero-shot prediction probability of class $c$ is then computed by the cosine similarity between the visual feature and the text prototypes:
\begin{equation}
p_c(x_i) = 
\frac{\exp(\cos(f_i, g_c)/\tau)}
{\sum_{j=1}^{C} \exp(\cos(f_i, g_j)/\tau)},
\label{eq:clip_prob}
\end{equation}
where $\tau$ is the temperature parameter.

\begin{figure*}[t]
    \centering
    \includegraphics[width=.85\linewidth]{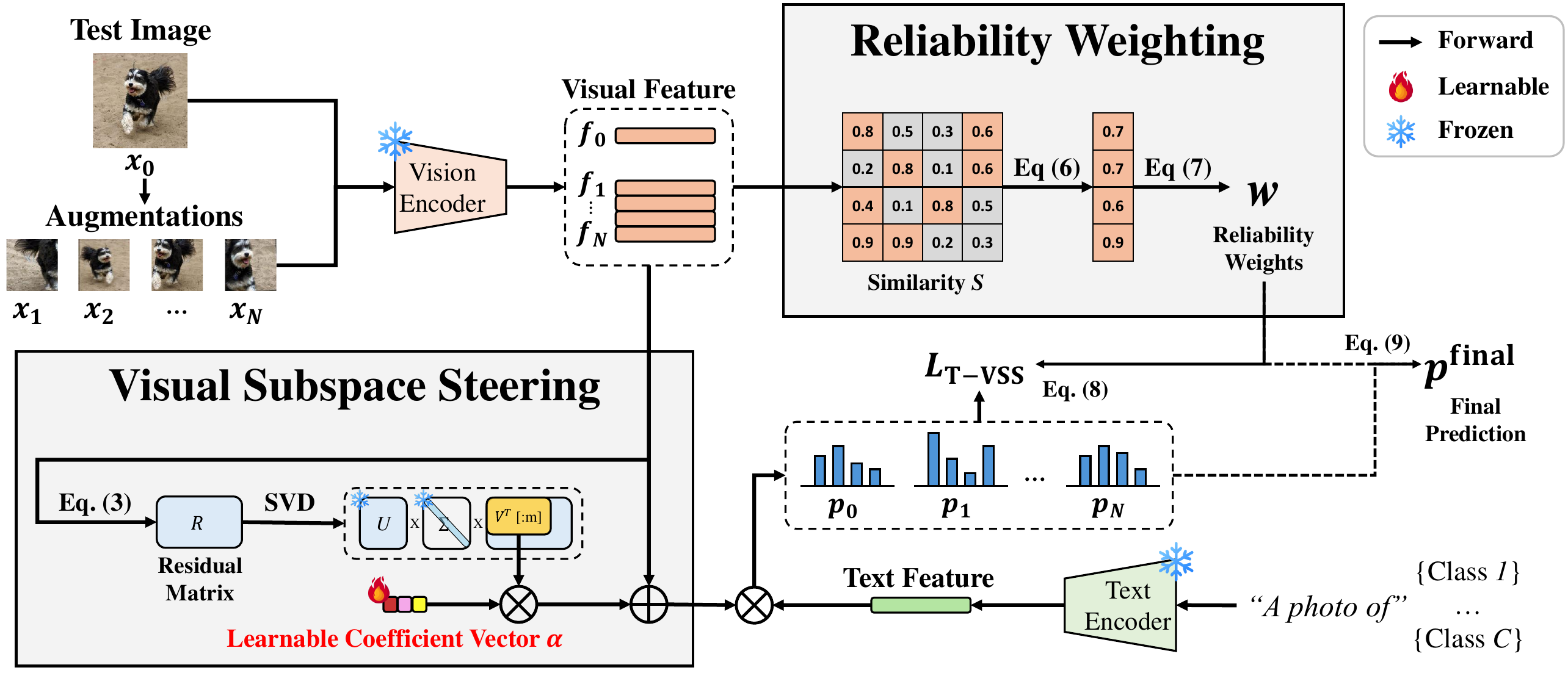}
    \vspace{-0.1cm}
    \caption{
Overview of \textbf{T-VSS}. 
From multi-view CLIP visual features, T-VSS applies Singular Value Decomposition (SVD) to anchor-based residuals to extract a compact visual subspace that captures dominant anchor-relative feature differences, then learns a shared low-rank correction within that subspace via reliability-weighted entropy minimization.
Predictions from the adapted views are finally aggregated using reliability-aware ensembling.
}
    \label{fig:method}
\vspace{-0.1cm}
\end{figure*}

\paragraph{Adversarial Test-Time Adaptation.}
Following prior test-time defenses~\cite{sheng2025r,ttp}, we construct $N+1$ stochastic views from a potentially adversarial test image $x$:
$\mathcal{X}(x)=\{x_0,x_1,\dots,x_N\}$, where $x_0=x$ and $\{x_i\}_{i=1}^N$ are augmented views of $x$.
In adversarial evaluation, all views are generated from the attacked image and therefore share the same perturbation source.
Let $p(x_i)\in\mathbb{R}^C$ denote the class probability vector in Eq.~\ref{eq:clip_prob}. 
We measure its prediction uncertainty with Shannon entropy:
\begin{equation}
\mathcal{H}(p(x_i)) = -\sum_{c=1}^{C} p_c(x_i)\log p_c(x_i).
\end{equation}

\subsection{Motivating Observations}
\label{sec:motivation}

We first examine the structure of attacked multi-view features. Using paired clean and adversarial images from the 50,000-image ImageNet~\cite{deng2009imagenet} validation set under the PGD~\cite{madry2018towards} attack settings, we extract features from one unaugmented original view and 63 stochastic views generated with matched augmentations for each clean--adversarial pair. Let $f_i^s$ denote the feature of view $i$ under setting $s\in\{\mathrm{clean},\mathrm{adv}\}$, where $i=0$ denotes the original view. We conduct two complementary analyses. First, we define the adversarial shift of each view as $\delta_i=f_i^{\mathrm{adv}}-f_i^{\mathrm{clean}}$ and measure the average cross-view cosine similarity and shared-energy ratio of these shifts. Second, treating $f_0^s$ as the anchor for each setting, we form residuals $r_i^s=f_i^s-f_0^s$ for the 63 stochastic views and
measure the minimum number of singular directions required to capture 90\% of their total squared energy.
Exact definitions are provided in Appendix~\ref{app:diagnostic_metrics}.

\begin{table}[t]
\centering
\caption{Adversarial-shift and anchor-relative residual statistics on the ImageNet validation set. Residual ranks capture 90\% of the anchor-relative
residual energy.}
\label{tab:paired_feature_analysis}
\resizebox{1\linewidth}{!}{%
\begin{tabular}{lcccc}
\toprule
\multirow{2}{*}{Backbone} & \multicolumn{2}{c}{\makecell{Adversarial Shift \\ Structure}} & \multicolumn{2}{c}{\makecell{Anchor-relative \\ Residual Rank}} \\
\cmidrule(lr){2-3}\cmidrule(lr){4-5}
& Cosine Similarity & Shared Energy & Clean & Adv. \\
\midrule
ResNet-50 & 0.314 & 0.230 & 8.28 & 3.37 \\
ViT-B/16 & 0.498 & 0.435 & 8.58 & 1.45 \\
ViT-B/32 & 0.475 & 0.435 & 10.31 & 2.13 \\
\bottomrule
\end{tabular}
}
\end{table}

Table~\ref{tab:paired_feature_analysis} shows two consistent patterns. The shift statistics indicate that adversarial effects remain partially aligned across views, motivating a common correction rather than independent view-wise updates.
Separately, a small number of singular directions captures most of the anchor-relative residual energy in both clean and adversarial settings, with stronger concentration under attack.
This observation motivates using the dominant residual directions as a compact search space for adaptation.

\subsection{Test-time Visual Subspace Steering}

Motivated by the cross-view shift alignment and residual energy concentration above, T-VSS learns a shared correction within a sample-specific low-rank visual subspace. The residual basis constrains where adaptation may move, while reliability-weighted entropy minimization determines the direction and magnitude of the correction. The image encoder and text prototypes remain fixed throughout; only a low-dimensional coefficient vector is optimized at test time.

\paragraph{Local Visual Subspace Estimation.}
We begin by extracting normalized CLIP visual features $\{f_i = F(x_i)\}_{i=0}^{N}$ for all views.
Instead of optimizing an unconstrained shift in the full $d$-dimensional embedding space,
T-VSS first estimates a compact subspace that captures dominant feature differences between the original view and its augmentations.
We use the original attacked view $x_0$ as an anchor and build the residual matrix
\begin{equation}
R = 
\begin{bmatrix}
(f_1 - f_0)^\top \\
(f_2 - f_0)^\top \\
\vdots \\
(f_N - f_0)^\top
\end{bmatrix}
\in \mathbb{R}^{N \times d}.
\label{eq:residual_matrix}
\end{equation}
We then compute the singular value decomposition $R = U\Sigma V^\top$, where the right singular vectors in $V$ define orthogonal directions in the visual embedding space.
Rather than fixing the rank manually, we choose the smallest active rank $m$ whose cumulative singular-value energy exceeds a threshold $\rho$:
\begin{equation}
m = \min \left\{ q \; \middle| \; 
\frac{\sum_{j=1}^{q}\sigma_j^2}{\sum_{j=1}^{\mathrm{rank}(R)}\sigma_j^2} \ge \rho
\right\},
\label{eq:rank_selection}
\end{equation}
where $\{\sigma_j\}$ are the singular values and $\rho \in (0,1]$ is the retained-energy threshold.
Anchoring at $f_0$ expresses each augmented feature relative to the original test view. These residuals capture both the displacement from the anchor to the augmented-feature mean and the variation around that mean.
Their leading right singular vectors define a compact, sample-specific search space for correction, rather than an estimate of the adversarial displacement itself.

\paragraph{Shared Subspace Steering.}
Following the cross-view alignment observed in Sec.~\ref{sec:motivation}, T-VSS learns one consensus correction rather than adapting each view independently. Given the basis $V_m \in \mathbb{R}^{d \times m}$, which collects the top-$m$ right singular vectors, we initialize a coefficient vector $\alpha \in \mathbb{R}^{m}$ at zero and generate the shared shift $\Delta = V_m \alpha$.
The same shift is then applied to every view as follows:
\begin{equation}
\tilde{f}_i = \frac{f_i + \Delta}{\|f_i + \Delta\|_2}, \quad i=0,\dots,N.
\label{eq:adapted_feature}
\end{equation}
Importantly, the residual-SVD basis determines only the admissible search space; reliability-weighted entropy minimization
selects the correction within it. Parameterizing $\Delta=V_m\alpha$ therefore restricts the update to the dominant directions of the anchor-relative residuals and avoids unconstrained optimization in the full embedding space. Each adapted representation is still formed from its view-specific feature $f_i$. The shared shift couples adaptation across views and discourages independent fitting to augmentation artifacts or spurious low-entropy directions.
Since the only learnable variable is the $m$-dimensional coefficient vector, the number of sample-wise trainable parameters is exactly the selected rank.

\paragraph{Reliability-Aware Optimization.}

Not all stochastic views are equally informative under attack. T-VSS therefore assigns each view a reliability score~\cite{sheng2025r} as a practical proxy for how well that view agrees with the local feature neighborhood. Because CLIP already outputs $\ell_2$-normalized image features, this agreement can be computed directly from pairwise similarities $S_{ij} = f_i^\top f_j$.
For each view, we average its top-$K$ nearest neighbor similarities:
\begin{equation}
r_i = \frac{1}{K}\sum_{j \in \mathcal{N}_K(i)} S_{ij},
\label{eq:topk}
\end{equation}
where $\mathcal{N}_K(i)$ denotes the indices of the top-$K$ most similar views excluding itself. We then convert these scores into reliability weights with a temperature-scaled softmax:
\begin{equation}
w_i = \frac{\exp(r_i/\tau_r)}{\sum_{j=0}^{N}\exp(r_j/\tau_r)},
\label{eq:reliability}
\end{equation}
where $\tau_r$ is a reliability temperature. Views that remain in a dense and mutually consistent feature neighborhood receive larger weights, while unstable outliers are suppressed.

Given the adapted features $\{\tilde{f}_i\}$, we compute logits with the frozen CLIP text prototypes and denote the resulting class probabilities by $p(\tilde{f}_i)$. We then optimize $\alpha$ by minimizing reliability-weighted pointwise entropy:
\begin{equation}
\mathcal{L}_{\mathrm{T-VSS}}
=
\sum_{i=0}^{N} w_i \, \mathcal{H}\bigl(p(\tilde{f}_i)\bigr),
\label{eq:vss_loss}
\end{equation}
where $w_i$ is the reliability weight defined in Eq.~\ref{eq:reliability}. 
In contrast to hard view selection, the loss softly uses all views while reducing the influence of unreliable ones.

\paragraph{Final Prediction and Efficiency.}
After optimization, the final prediction is obtained by reliability-weighted averaging of the adapted view probabilities:
\begin{equation}
p^{\mathrm{final}} = \sum_{i=0}^{N} w_i \, p(\tilde{f}_i),
\qquad
\hat{y} = \arg\max_{c} \; p^{\mathrm{final}}_c.
\label{eq:final_pred}
\end{equation}
T-VSS encodes the image views only once and performs all subsequent optimization directly on cached visual features.
Unlike prompt- or pixel-space methods, it avoids repeated backpropagation through large model components or dense input variables.
Overall, T-VSS remains lightweight and easy to interpret as a sample-wise low-rank correction of the attacked visual representation guided by multi-view agreement.

\section{Experiments}

\subsection{Setup}
\label{exp:setup}
\paragraph{Datasets and Models.}

We evaluate T-VSS on both fine-grained recognition benchmarks and large-scale ImageNet-style benchmarks. For fine-grained evaluation, we use eight datasets spanning diverse visual domains: Caltech101~\cite{fei2004learning}, Pets~\cite{parkhi2012cats}, Flower102~\cite{nilsback2008automated}, Stanford Cars~\cite{krause20133d}, FGVC Aircraft~\cite{maji2013fine}, DTD~\cite{cimpoi2014describing}, EuroSAT~\cite{helber2019eurosat}, and UCF101~\cite{soomro2012ucf101}. We further evaluate on ImageNet~\cite{deng2009imagenet} and four ImageNet out-of-distribution benchmarks: ImageNet-A~\cite{hendrycks2021natural}, ImageNet-V2~\cite{recht2019imagenet}, ImageNet-R~\cite{hendrycks2021many}, and ImageNet-S~\cite{wang2019learning}. As the underlying VLM, we adopt official CLIP checkpoints and consider three widely used backbones: ResNet-50, ViT-B/16, and ViT-L/14.

\paragraph{Evaluation and Baselines.}
We report both clean top-1 accuracy (\textbf{Acc.}) and adversarial top-1 accuracy (\textbf{Rob.}). Following prior work on adversarial test-time defense for CLIP, adversarial examples are generated under a white-box but defense-oblivious threat model against the original CLIP model using PGD, while the test-time defense remains hidden from the attacker. We compare T-VSS with vanilla CLIP, a simple multi-view \textbf{Ensemble}, standard VLM test-time adaptation baselines, and recent adversarial test-time defenses. Depending on the benchmark and backbone, the comparison set includes \textbf{TPT}~\cite{shu2022test}, \textbf{C-TPT}~\cite{yoon2024c}, \textbf{MTA}~\cite{zanella2024test}, \textbf{TTC}~\cite{xing2025clip}, \textbf{ET3}~\cite{ET3}, \textbf{R-TPT}~\cite{sheng2025r}, and \textbf{TTP}~\cite{ttp}.

\paragraph{Implementation Details.}
For adversarial evaluation, we generate $\ell_\infty$-bounded PGD~\cite{madry2018towards} examples with backbone-specific settings following standard CLIP robustness benchmarks. For ResNet-50, we use a perturbation budget of $\epsilon = 1/255$ and 7 attack steps. For ViT-B/16 and ViT-L/14, we use a stronger setting with $\epsilon = 4/255$ and 100 attack steps.
In all cases, the attack step size is set to $\epsilon/4$.
For zero-shot classification, we use the default hand-crafted prompt template ``a photo of a [CLASS]'' to construct text prototypes. We perform a single update step using AdamW with a learning rate of $0.1$. We set the retained-energy threshold $\rho$ to 0.9, construct the visual subspace with anchor-based residuals, and compute reliability weights with temperature $0.05$ using a default top-$K$ neighbor count of $K=5$.
Each test sample is processed with 64 views in total, including the original image and 63 augmented views generated using AugMix~\cite{hendrycks2020augmix}.

\begin{table*}[ht!]
\centering
\caption{Clean (Acc.) and adversarial (Rob.) top-1 accuracy (\%) on eight fine-grained datasets across three CLIP backbones. Best clean accuracy and best adversarial accuracy are highlighted in \textbf{bold} and \textbf{\textcolor{blush}{bold}}, respectively. $\dagger$ indicates reproduced results.
}
\vspace{-0.2cm}
\label{tab:combined_finegrained_backbones}
\resizebox{.98\linewidth}{!}{%
\begin{tabular}{lcacacacacacacacaca}
\toprule
\multicolumn{1}{l}{Method} 
& \multicolumn{2}{c}{Caltech101} 
& \multicolumn{2}{c}{Pets} 
& \multicolumn{2}{c}{Cars} 
& \multicolumn{2}{c}{Flower102} 
& \multicolumn{2}{c}{Aircraft} 
& \multicolumn{2}{c}{DTD} 
& \multicolumn{2}{c}{EuroSAT} 
& \multicolumn{2}{c}{UCF101} 
& \multicolumn{2}{c}{Avg.} \\
\cmidrule(lr){2-3}\cmidrule(lr){4-5}\cmidrule(lr){6-7}\cmidrule(lr){8-9}\cmidrule(lr){10-11}\cmidrule(lr){12-13}\cmidrule(lr){14-15}\cmidrule(lr){16-17}\cmidrule(lr){18-19}
\multicolumn{1}{c}{} 
& \multicolumn{1}{c}{Acc.} & \multicolumn{1}{c}{Rob.} 
& \multicolumn{1}{c}{Acc.} & \multicolumn{1}{c}{Rob.} 
& \multicolumn{1}{c}{Acc.} & \multicolumn{1}{c}{Rob.} 
& \multicolumn{1}{c}{Acc.} & \multicolumn{1}{c}{Rob.} 
& \multicolumn{1}{c}{Acc.} & \multicolumn{1}{c}{Rob.} 
& \multicolumn{1}{c}{Acc.} & \multicolumn{1}{c}{Rob.} 
& \multicolumn{1}{c}{Acc.} & \multicolumn{1}{c}{Rob.} 
& \multicolumn{1}{c}{Acc.} & \multicolumn{1}{c}{Rob.} 
& \multicolumn{1}{c}{Acc.} & \multicolumn{1}{c}{Rob.} \\
\midrule

\multicolumn{19}{c}{\textbf{CLIP-ResNet-50} ($\epsilon=1/255$)} \\
\midrule
CLIP & 85.9 & 2.6 & 83.5 & 0.0 & 55.7 & 0.0 & 61.7 & 0.0 & 15.7 & 0.0 & 40.4 & 0.8 & 23.7 & 0.0 & 58.9 & 0.0 & 53.2 & 0.4 \\
ET3\dag & 79.5 & 64.6 & 77.6 & 38.2 & 47.4 & 18.3 & 53.6 & 18.9 & 11.4 & 5.0 & 34.3 & 21.7 & 11.8 & 11.4 & 50.2 & 31.4 & 45.7 & 26.2 \\
Ensemble & 83.5 & 74.8 & 82.3 & 69.9 & 57.1 & 36.2 & 58.0 & 46.6 & 16.4 & 9.8 & 37.1 & 29.5 & 16.7 & 13.7 & 53.9 & 43.0 & 50.6 & 40.4 \\
TPT & \bestclean{87.9} & 7.0 & 84.7 & 0.1 & 58.4 & 0.0 & 62.1 & 0.0 & 17.3 & 0.0 & \bestclean{42.4} & 4.3 & \bestclean{28.4} & 0.0 & \bestclean{60.6} & 0.3 & \bestclean{55.2} & 1.5 \\
C-TPT & 87.7 & 3.7 & 83.6 & 0.0 & 56.6 & 0.0 & \bestclean{64.8} & 0.0 & 16.7 & 0.0 & 41.5 & 1.3 & 27.0 & 0.0 & 60.1 & 0.1 & 54.8 & 0.6 \\
MTA & 87.3 & 65.9 & \bestclean{84.8} & 59.8 & \bestclean{58.7} & 17.8 & 61.0 & 31.5 & \bestclean{18.1} & 3.7 & 40.3 & 18.8 & 22.5 & 1.6 & \bestclean{60.6} & 31.3 & 54.1 & 28.8 \\
R-TPT & 86.7 & \bestrob{79.8} & 84.6 & 74.2 & 58.1 & 42.9 & 60.6 & 51.9 & 17.5 & 12.6 & 41.3 & 33.5 & 21.2 & 15.9 & 59.7 & 50.9 & 53.7 & 45.2 \\
TTP\dag & 85.9 & 73.7 & 83.8 & 46.9 & 55.6 & 38.9 & 61.0 & 35.2 & 15.6 & 11.6 & 40.1 & 25.9 & 24.0 & \bestrob{23.4} & 58.5 & 49.0 & 53.1 & 38.1 \\
T-VSS (Ours) & 84.4 & 78.1 & 84.4 & \bestrob{75.5} & 55.6 & \bestrob{54.7} & 58.8 & \bestrob{54.0} & 18.0 & \bestrob{20.3} & 38.9 & \bestrob{35.5} & 18.5 & 17.1 & 58.9 & \bestrob{52.5} & 52.2 & \bestrob{48.5} \\
\midrule

\multicolumn{19}{c}{\textbf{CLIP-ViT-B/16} ($\epsilon=4/255$)} \\
\midrule
CLIP & 94.0 & 0.0 & \bestclean{88.3} & 0.0 & 65.5 & 0.0 & 67.4 & 0.0 & 23.9 & 0.0 & 44.4 & 0.0 & 42.2 & 0.0 & 65.2 & 0.0 & 61.4 & 0.0 \\
TTC & 87.6 & 8.4 & 82.3 & 10.4 & 55.0 & 2.9 & \bestclean{69.0} & 7.4 & 23.3 & 0.5 & 41.0 & 4.5 & \bestclean{47.4} & 0.4 & 65.8 & 1.6 & 58.9 & 4.5 \\
ET3\dag & 88.8 & 17.5 & 82.2 & 14.0 & 57.4 & 0.8 & 62.9 & 3.5 & 18.9 & 0.4 & 38.8 & 5.0 & 31.5 & 0.8 & 57.8 & 2.4 & 54.8 & 5.6
\\
Ensemble & 91.9 & 74.7 & 86.2 & 51.2 & 65.7 & 26.0 & 65.9 & 36.3 & 23.4 & 8.7 & 43.2 & 25.1 & 28.2 & 2.2 & 63.0 & 30.6 & 58.4 & 31.8 \\
MTA & \bestclean{94.3} & 72.1 & 88.0 & 51.8 & \bestclean{67.7} & 18.5 & 67.4 & 27.9 & \bestclean{25.0} & 4.3 & \bestclean{46.5} & 16.2 & 42.5 & 1.2 & \bestclean{67.5} & 27.5 & \bestclean{62.3} & 27.4 \\
R-TPT & 93.7 & 82.0 & 87.2 & 60.2 & 67.0 & 34.7 & 68.7 & 44.6 & 23.9 & 13.2 & 46.4 & 32.8 & 34.7 & 8.5 & 67.2 & 43.2 & 61.1 & 39.9 \\
TTP & 93.5 & \bestrob{82.3} & \bestclean{88.3} & 64.7 & 65.4 & 37.4 & 67.3 & 47.2 & 23.9 & 14.8 & 44.1 & 36.0 & 42.0 & \bestrob{14.5} & 65.0 & \bestrob{47.2} & 61.2 & 42.9 \\
T-VSS (Ours) & 93.4 & 81.5 & 87.3 & \bestrob{64.9} & 65.8 & \bestrob{54.5} & 65.9 & \bestrob{50.6} & 24.3 & \bestrob{24.4} & 45.9 & \bestrob{37.4} & 34.8 & 7.5 & 66.0 & 45.2 & 60.4 & \bestrob{45.8} \\
\midrule

\multicolumn{19}{c}{\textbf{CLIP-ViT-L/14} ($\epsilon=4/255$)} \\
\midrule
CLIP & 95.2 & 0.1 & 93.1 & 0.0 & 76.8 & 0.0 & 76.2 & 0.0 & 30.0 & 0.0 & 52.4 & 0.0 & 55.1 & 0.0 & 73.7 & 0.0 & \bestclean{69.1} & 0.0 \\
TTC & 88.7 & 7.7 & 92.2 & 7.6 & 67.8 & 2.2 & \bestclean{76.5} & 7.5 & 31.7 & 0.5 & 49.7 & 6.2 & \bestclean{64.1} & 0.2 & \bestclean{75.0} & 2.2 & 68.2 & 4.3 \\
ET3\dag & 92.5 & 24.1 & 89.9 & 19.3 & 72.6 & 1.5 & 73.7 & 7.5 & 28.5 & 0.6 & 46.0 & 7.4 & 39.6 & 4.0 & 71.1 & 6.3 & 64.2 & 8.8
\\
Ensemble & 94.9 & 83.6 & 93.4 & 63.5 & 76.3 & 40.5 & 75.0 & 48.6 & 31.7 & 12.7 & 51.3 & 31.3 & 38.7 & 11.1 & 71.7 & 48.3 & 66.6 & 42.5 \\
MTA & \bestclean{95.8} & 83.1 & \bestclean{93.7} & 64.9 & \bestclean{78.4} & 36.6 & 76.1 & 44.2 & \bestclean{32.7} & 8.0 & 53.4 & 27.2 & 47.8 & 7.5 & 74.7 & 47.5 & \bestclean{69.1} & 39.9 \\
R-TPT & 95.7 & 88.2 & \bestclean{93.7} & 72.9 & 77.2 & 49.1 & 76.2 & 55.6 & 31.7 & 17.2 & \bestclean{54.0} & 38.0 & 44.3 & 20.4 & 74.3 & 55.6 & 68.4 & 49.6 \\
TTP & 95.1 & \bestrob{88.6} & 93.1 & \bestrob{76.3} & 76.8 & 51.1 & 76.1 & 58.7 & 29.2 & 17.7 & 52.3 & 41.3 & 55.0 & \bestrob{21.6} & 73.6 & \bestrob{57.4} & 68.9 & 51.6 \\
T-VSS (Ours) & 94.8 & 87.5 & \bestclean{93.7} & 73.8 & 76.3 & \bestrob{63.2} & 75.2 & \bestrob{60.9} & \bestclean{32.7} & \bestrob{26.9} & 53.4 & \bestrob{41.7} & 45.1 & 20.3 & 73.5 & 55.9 & 68.1 & \bestrob{53.8} \\
\bottomrule
\end{tabular}
}
\vspace{-0.2cm}
\end{table*}

\begin{table*}[ht!]

\begin{minipage}[t]{0.68\textwidth}
  \centering
  \caption{Clean (Acc.) and adversarial (Rob.) top-1 accuracy (\%) on ImageNet and four ImageNet-OOD benchmarks with CLIP-ResNet-50. $\dagger$ indicates reproduced results.}
  \vspace{-0.2cm}
  \label{tab:imagenet}
  \resizebox{0.97\linewidth}{!}{%
  \begin{tabular}{lcacacacacaca}
    \toprule
    \multirow{2}{*}{Method} & \multicolumn{2}{c}{ImageNet} & \multicolumn{2}{c}{ImageNet-A} & \multicolumn{2}{c}{ImageNet-V2} & \multicolumn{2}{c}{ImageNet-R} & \multicolumn{2}{c}{ImageNet-S} & \multicolumn{2}{c}{Avg.} \\
    \cmidrule(lr){2-3}\cmidrule(lr){4-5}\cmidrule(lr){6-7}\cmidrule(lr){8-9}\cmidrule(lr){10-11}\cmidrule(lr){12-13}
    & Acc. & \mc{1}{Rob.} & Acc. & \mc{1}{Rob.} & Acc. & \mc{1}{Rob.} & Acc. & \mc{1}{Rob.} & Acc. & \mc{1}{Rob.} & Acc. & \mc{1}{Rob.} \\
    \midrule
    CLIP & 58.2 & 0.1 & 21.8 & 0.0 & 51.5 & 0.1 & 56.1 & 0.8 & 33.3 & 0.5 & 44.2 & 0.3 \\
    ET3\dag & 52.0 & 22.0 & 12.4 & 3.1 & 45.5 & 18.8 & 50.0 & 27.6 & 29.1 & 21.2 & 37.8 & 18.5
\\
    Ensemble & 58.0 & 40.1 & 22.6 & 10.1 & 52.0 & 37.2 & 51.3 & 39.3 & 29.5 & 20.7 & 42.7 & 29.5 \\
    TPT & 60.7 & 0.3 & 26.5 & 0.0 & 54.8 & 0.3 & \bestclean{58.9} & 1.8 & 35.0 & 1.4 & \bestclean{47.2} & 0.7 \\
    C-TPT  & 60.4 & 0.1 & 24.1 & 0.0 & 54.3 & 0.1 & 57.7 & 1.0 & 34.7 & 0.9 & 46.2 & 0.4 \\
    MTA  & 60.4 & 30.0 & 27.5 & 5.6 & 54.2 & 24.6 & 58.4 & 29.8 & \bestclean{35.2} & 11.3 & 47.1 & 20.3 \\
    R-TPT & \bestclean{60.9} & 47.7 & \bestclean{28.4} & 14.4 & \bestclean{54.9} & 41.6 & 57.6 & 46.9 & 34.0 & 26.2 & 47.1 & 35.4 \\
    TTP\dag & 58.2 & 43.0 & 21.9 & 12.6 & 51.3 & 38.1 & 56.1 & 38.3 & 33.4 & 17.3 & 44.2 & 29.9 \\
    T-VSS (Ours) & 59.7 & \bestrob{50.1} & 27.8 & \bestrob{16.0} & 53.6 & \bestrob{44.6} & 57.2 & \bestrob{47.6} & 33.6 & \bestrob{29.8} & 46.4 & \bestrob{37.6} \\
    \bottomrule
  \end{tabular}%
  }

    \end{minipage}
    \hfill
    \begin{minipage}[t]{0.31\textwidth}
            \centering
            \caption{Per-image latency and adversarial accuracy (\%) on UCF101 with CLIP-ResNet-50 under different view budgets.}
            \label{tab:efficiency_ucf101}
            \resizebox{0.99\columnwidth}{!}{%
            \begin{tabular}{lca}
            \toprule
            Method & \makecell{Running time \\ (s/image)} & \multicolumn{1}{c}{Rob.} \\
            \midrule
            R-TPT (64 views) & 0.533 & 50.9 \\
            TTP (64 views)   & 0.408 & 49.0 \\
            \midrule
            T-VSS (64 views) & 0.383 & \bestrob{52.5} \\
            T-VSS (32 views) & 0.193 & 52.2 \\
            T-VSS (16 views) & 0.092 & 51.1 \\
            T-VSS (8 views) & 0.046 & 49.0 \\
            \bottomrule
            \end{tabular}
            }
    \end{minipage}
    \vspace{-0.2cm}
\end{table*}

\subsection{Experimental Results}

\paragraph{Results on Fine-grained Datasets.}

Table~\ref{tab:combined_finegrained_backbones} reports clean and adversarial accuracy on eight fine-grained datasets across three CLIP backbones. T-VSS achieves the best average robust accuracy on all three backbones, reaching 48.5\% on ResNet-50, 45.8\% on ViT-B/16, and 53.8\% on ViT-L/14.
These scores exceed those of the strongest prior defense by 3.3, 2.9, and 2.2 points, respectively, while keeping clean accuracy competitive.
Overall, the results show that the benefit of T-VSS is not confined to a particular backbone or dataset, but extends consistently across architectures while improving the robustness--accuracy trade-off.
Notably, the margin is especially clear on ResNet-50, where padding-based TTP is less competitive than on ViT backbones. This pattern suggests that input-space padding may transfer less reliably across backbone families than feature-space correction, since padding operates at the pixel-level while T-VSS adapts visual representations directly.

The per-dataset results further clarify where direct feature correction is most beneficial. T-VSS is especially strong on challenging fine-grained datasets such as Cars, Flower102, Aircraft, and DTD, where it achieves the best robust accuracy on all three backbones. It also remains competitive on UCF101, although TTP is slightly stronger on the two larger ViT backbones.
EuroSAT is the clearest exception, where TTP achieves higher robust accuracy on all three backbones, indicating that the advantage of T-VSS varies across datasets.
Despite this exception, T-VSS remains the strongest method on average in terms of robust accuracy across all three backbones.
Additional results are in Tables~\ref{tab:tecoa_pretrained} and~\ref{tab:vitb32}.

\paragraph{Results on ImageNet and ImageNet-OOD Datasets.}
Table~\ref{tab:imagenet} shows that the gains of T-VSS extend beyond fine-grained recognition to large-scale and out-of-distribution evaluation on CLIP-ResNet-50. T-VSS obtains the best robust accuracy on ImageNet and on every OOD benchmark, improving the average robust accuracy to 37.6\% and surpassing the previous best defense, R-TPT, by 2.2 points. The gain is consistent across ImageNet-A, ImageNet-V2, ImageNet-R, and ImageNet-S, which suggests that the proposed feature-space correction is not tied to a specific dataset bias or shift type.
T-VSS also preserves competitive clean accuracy. Although some prompt-based methods obtain slightly higher clean scores, their adversarial robustness remains substantially lower. This comparison highlights the central advantage of T-VSS: by adapting the attacked visual representation directly, it improves robustness without paying the large clean-accuracy penalty often associated with aggressive test-time correction. Taken together with the fine-grained results, these experiments support T-VSS as a robust test-time defense for zero-shot VLM inference.

\section{Analysis and Ablation}
\label{sec:analysis}

\paragraph{Analysis of Inference Efficiency.}
Table~\ref{tab:efficiency_ucf101} compares per-image latency
and adversarial accuracy on UCF101 with CLIP-ResNet-50.
Under the same 64-view budget, T-VSS achieves the lowest
latency among the compared methods at 0.383 seconds per
image, while attaining the highest robust accuracy of 52.5\%.
T-VSS encodes the views once and performs adaptation
directly on cached features using low-dimensional coefficients.
With 16 views, latency decreases to 0.092 seconds per image,
providing a $5.8\times$ speedup over 64-view R-TPT while
retaining higher robust accuracy (51.1\% vs.\ 50.9\%).

\begin{figure}[t]
    \centering
    \includegraphics[width=.8\linewidth]{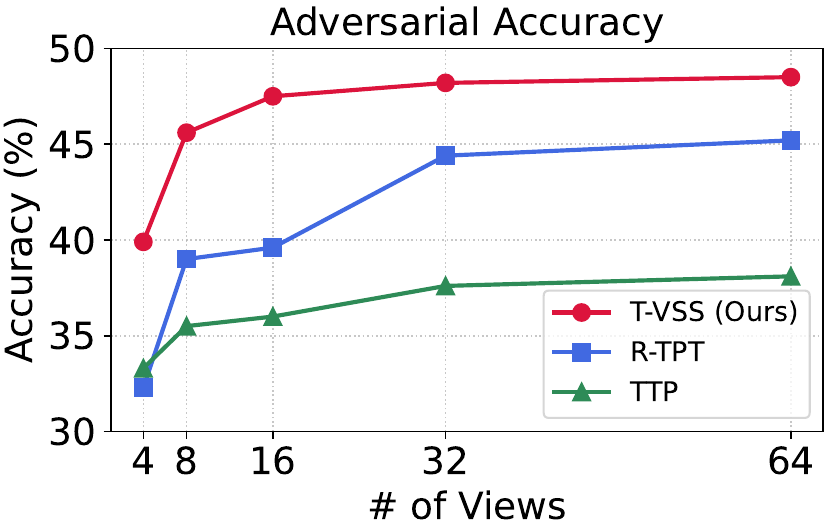}
    \vspace{-0.2cm}
    \caption{
    Ablation of the number of views.
    }
    \vspace{-0.3cm}
    \label{fig:view_ablation_rob}
\end{figure}

\paragraph{Robustness under Different View Budgets.}
Figure~\ref{fig:view_ablation_rob} reports average adversarial
accuracy across eight fine-grained datasets with CLIP-ResNet-50.
T-VSS achieves the highest accuracy at every evaluated
view budget, with a particularly clear advantage in the
low-view regime.
With only 8 views, it already rivals 64-view R-TPT
and outperforms 64-view TTP, demonstrating effective
adaptation under limited augmentation budgets.
Such view efficiency is valuable when test-time latency or augmentation budget is limited.

\begin{table}[t]
    \centering
    \caption{Adversarial accuracy (\%) under additional attacks on Flower102 and DTD using CLIP-ViT-B/16. DF denotes DeepFool.}
    \vspace{-0.2cm}
    \label{tab:other-attack}
    \resizebox{1\columnwidth}{!}{%
    \begin{tabular}{laaaaaaaa}
    \toprule
    \multirow{2}{*}{Method} & \multicolumn{4}{c}{Flower102} & \multicolumn{4}{c}{DTD} \\
    \cmidrule(lr){2-5}\cmidrule(lr){6-9}
     & \multicolumn{1}{c}{CW} & \multicolumn{1}{c}{DF} & \multicolumn{1}{c}{FGSM} & \multicolumn{1}{c}{Avg.} & \multicolumn{1}{c}{CW} & \multicolumn{1}{c}{DF} & \multicolumn{1}{c}{FGSM} & \multicolumn{1}{c}{Avg.} \\
    \midrule
    CLIP & 0.8  & 0.4  & 4.8  & 2.0  & 2.3  & 7.6  & 13.4 & 7.8 \\
    Ensemble & 50.1 & 52.2 & 46.6 & 49.7 & 31.1 & 32.9 & 29.7 & 31.2 \\
    TPT & 13.8 & 10.8 & 14.2 & 12.9 & 21.3 & 24.4 & 22.2 & 22.6 \\
    C-TPT & 6.6  & 5.5  & 6.2  & 6.1  & 11.9 & 15.8 & 17.5 & 15.1 \\
    MTA & 34.5 & 35.4 & 36.6 & 35.5 & 23.6 & 23.5 & 23.9 & 23.7 \\
    R-TPT & 51.6 & 54.7 & 49.2 & 51.8 & 34.2 & 35.9 & 32.5 & 34.2 \\
    TTP & 54.1 & 56.4 & 51.8 & 54.1 & 38.9 & 40.1 & 37.1 & 38.7 \\
    T-VSS (Ours) & \bestrob{54.8} & \bestrob{60.5} & \bestrob{53.7} & \bestrob{56.3} & \bestrob{39.1} & \bestrob{42.3} & \bestrob{37.4} & \bestrob{39.6} \\
    \bottomrule
    \end{tabular}
    }
    \vspace{-0.15cm}
\end{table}

\paragraph{Robustness under Various Attacks.}

Table~\ref{tab:other-attack} evaluates T-VSS under three additional attacks: $\ell_2$-based CW~\cite{carlini2017evaluating}, $\ell_2$-based DeepFool~\cite{moosavi2016deepfool}, and $\ell_\infty$-bounded FGSM~\cite{goodfellow2015explaining}, on Flower102 and DTD. T-VSS achieves the best adversarial accuracy across all attack settings and both datasets, reaching 56.3\% average robustness on Flower102 and 39.6\% on DTD.
The consistent improvements across the evaluated $\ell_\infty$- and $\ell_2$-based attacks suggest that T-VSS is not narrowly tied to the primary $\ell_\infty$-PGD setting.
Additional robustness results under stronger attacks are in Table~\ref{tab:additional_attacks_vitb16}.

\begin{table}[t]
    \centering
    \caption{Ablation of adaptive rank and reliability weighting.}
    \vspace{-0.2cm}
    \label{tab:ablation_rank_weight}
    \resizebox{.81\columnwidth}{!}{%
    \begin{tabular}{cccaca}
    \toprule
    \multirow{2}{*}{\makecell{Adaptive\\Rank}} & \multirow{2}{*}{\makecell{Reliability\\Weighting}} & \multicolumn{2}{c}{ResNet-50} & \multicolumn{2}{c}{ViT-B/16} \\
    \cmidrule(lr){3-4}\cmidrule(lr){5-6}
    & & Acc. & \mc{1}{Rob.} & Acc. & \mc{1}{Rob.} \\
    \midrule
     {\color{gray!25}\ding{55}} & {\color{gray!35}\ding{55}} & 50.6 & 44.1 & 59.2 & 38.4 \\
     {\color{gray!25}\ding{55}} & \checkmark & 51.2 & 44.5 & 59.6 & 38.5 \\
    \checkmark & {\color{gray!25}\ding{55}} & 51.5 & 48.1 & 59.9 & 45.7 \\
    \checkmark & \checkmark & \bestclean{52.2} & \bestrob{48.5} & \bestclean{60.4} & \bestrob{45.8} \\
    \bottomrule
    \end{tabular}
    }
    \vspace{-0.3cm}
\end{table}

\paragraph{Ablation of T-VSS Components.}

Table~\ref{tab:ablation_rank_weight} isolates adaptive rank selection and reliability weighting. Without adaptive rank selection, steering uses the full residual rank, retaining low-energy directions alongside dominant ones. Enabling rank selection alone improves robust accuracy from $44.1\%$ to $48.1\%$ on ResNet-50 and from $38.4\%$ to $45.7\%$ on ViT-B/16. This substantial gain supports restricting entropy minimization to the dominant residual directions. Reliability weighting emphasizes views with stronger feature agreement during both adaptation and aggregation. Adding it to adaptive rank selection further improves clean and robust accuracy on both backbones, with robust accuracy reaching $48.5\%$ and $45.8\%$, respectively. 
\begin{figure*}[ht!]
    \centering
    \begin{subfigure}[b]{0.32\textwidth}
        \centering
        \includegraphics[width=\linewidth]{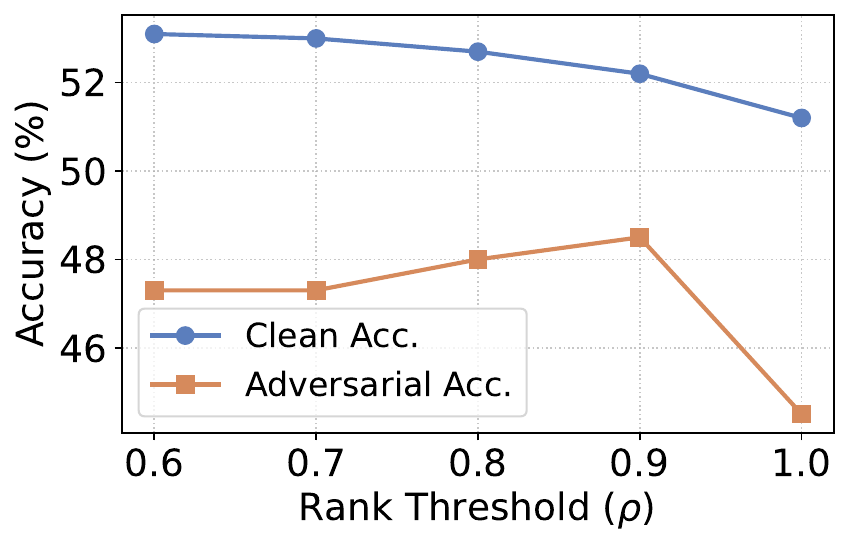}
        \caption{Rank Threshold $\rho$}
        \label{fig:rank_p}
    \end{subfigure}
    \hfill
    \begin{subfigure}[b]{0.32\textwidth}
        \centering
        \includegraphics[width=\linewidth]{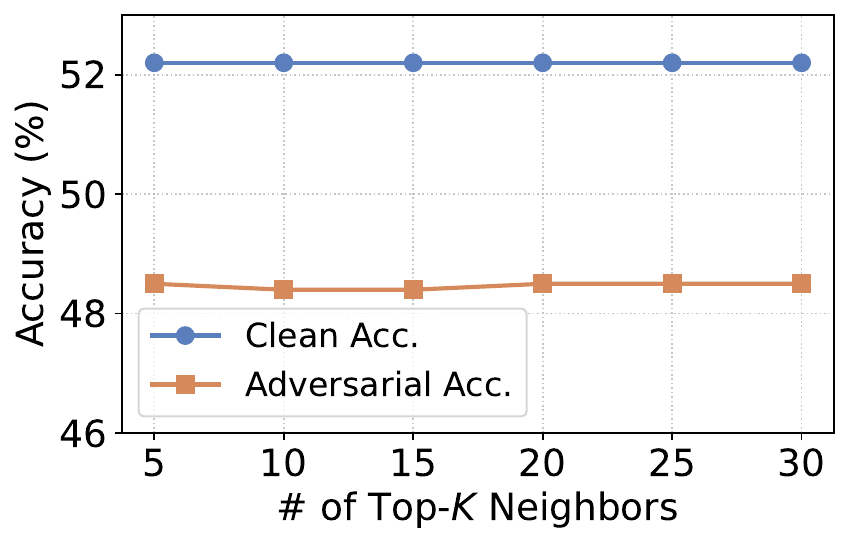}
        \caption{Top-$K$ Nearest Neighbor}
        \label{fig:top_k}
    \end{subfigure}
    \hfill
    \begin{subfigure}[b]{0.32\textwidth}
        \centering
        \includegraphics[width=\linewidth]{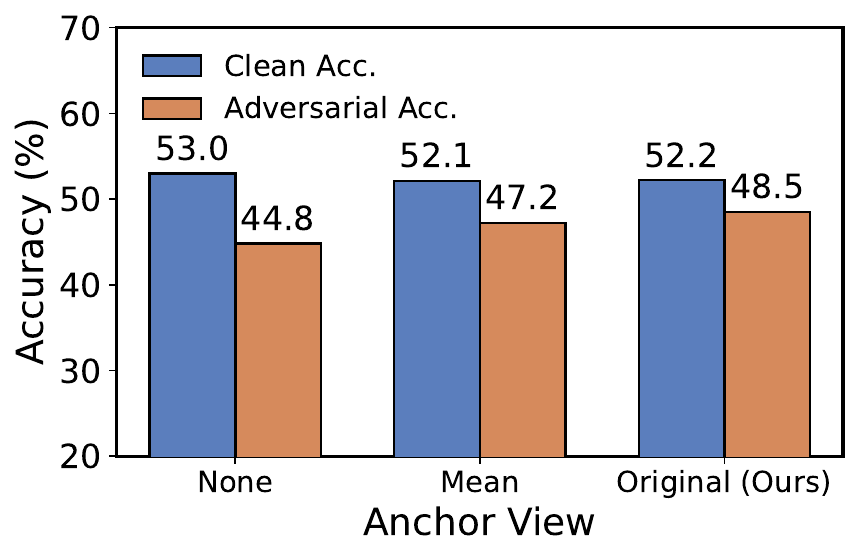}
        \caption{Anchor View}
        \label{fig:reference}
    \end{subfigure}
    \vspace{-0.2cm}
    
    \caption{Sensitivity of T-VSS to the rank threshold $\rho$, the number of neighbors $K$, and the choice of anchor view. 
    Results are averaged over the eight fine-grained datasets on CLIP-ResNet-50.
        }
        \label{fig:main_ablation}
    \vspace{-0.1cm}
\end{figure*}

\begin{table*}[h!]
\centering

\begin{minipage}[t]{0.32\textwidth}
    \vspace{0pt}
    \centering
    \caption{Comparison of residual-SVD and rank-matched random orthonormal bases.}
    \label{tab:random_basis_ablation}
    \resizebox{\linewidth}{!}{%
    \begin{tabular}{lcaca}
        \toprule
        Method
        & \multicolumn{2}{c}{ResNet-50}
        & \multicolumn{2}{c}{ViT-B/16}\\
        \cmidrule(lr){2-3}
        \cmidrule(lr){4-5}
        & Acc. & \mc{1}{Rob.}
        & Acc. & \mc{1}{Rob.}\\
        \midrule
        Random
        & 51.6 & 44.0
        & 59.3 & 38.2\\
        Residual-SVD (Ours)
        & \bestclean{52.2} & \bestrob{48.5}
        & \bestclean{60.4} & \bestrob{45.8}\\
        \bottomrule
    \end{tabular}%
    }
\end{minipage}
\hfill
\begin{minipage}[t]{0.15\textwidth}
    \vspace{0pt}
    \centering
    \caption{Comparison of per-view and shared steering.}
    \label{tab:shared_per_view}
    \resizebox{.99\linewidth}{!}{%
    \begin{tabular}{la}
        \toprule
        Correction & \mc{1}{Rob.} \\
        \midrule
        Per-view & 44.9 \\
        Shared (Ours) & \bestrob{45.8} \\
        \bottomrule
    \end{tabular}%
    }
\end{minipage}
\hfill
\begin{minipage}[t]{0.49\textwidth}
    \vspace{0pt}
    \centering
    \caption{Post-hoc analysis of the residual subspace. Before/After denote results without/with the learned feature shift. Clean alignment is the cosine similarity to the paired clean original-view feature.
    }
    \vspace{-0.15cm}
    \label{tab:steering_analysis}
    \resizebox{.85\linewidth}{!}{%
    \begin{tabular}{lcac}
        \toprule
        Backbone
        & \shortstack{Margin coverage (\%)\\Random / Residual}
        & \mc{1}{\shortstack{Rob. \\Before / After}}
        & \shortstack{Clean alignment\\Before / After} \\
        \midrule
        ResNet-50
        & 0.56 / \textbf{7.35}
        & 44.6 / \bestrob{48.5}
        & 0.731 / \textbf{0.745} \\
        ViT-B/16
        & 0.35 / \textbf{17.34}
        & 38.4 / \bestrob{45.8}
        & 0.347 / \textbf{0.374} \\
        \bottomrule
    \end{tabular}%
    }
\end{minipage}

\end{table*}

\paragraph{Ablation of Hyperparameters and Design Choices.}

Figure~\ref{fig:main_ablation} shows that T-VSS is stable across a broad range of hyperparameters.
For the rank threshold, robustness peaks near the default $\rho=0.9$, while the evaluated smaller thresholds still outperform R-TPT (45.2\% in Table~\ref{tab:combined_finegrained_backbones}). In contrast, retaining the full residual rank at $\rho=1.0$ substantially degrades performance, indicating that robustness depends on maintaining a compact subspace rather than finely tuning its exact rank.
Varying the top-$K$ neighborhood size used for reliability estimation in Eq.~\ref{eq:topk} has little effect on clean or robust accuracy, indicating low sensitivity to this hyperparameter. 
Finally, using the original test image as the anchor in Eq.~\ref{eq:residual_matrix} yields higher adversarial accuracy than both mean-centering and using raw features.

\paragraph{Importance of the Residual-SVD Basis.}

To distinguish the effect of residual geometry from generic low-dimensional restriction, we replace the residual-SVD basis with a random orthonormal basis of the same rank $m$, while keeping the pipeline unchanged. Table~\ref{tab:random_basis_ablation} shows that residual-SVD improves robust accuracy by $4.5$ points on ResNet-50 and $7.6$ points on ViT-B/16, with slightly higher clean accuracy.
These results show that the benefit cannot be explained by dimensionality reduction alone: at the same rank, directions extracted from anchor-relative residuals support more effective adaptation than random orthonormal directions.

\paragraph{Shared versus Per-view Feature Steering.}

The cross-view alignment in Sec.~\ref{sec:motivation} motivates a shared correction. We compare it with per-view steering, which learns a separate coefficient vector $\alpha_i$ for each view, yielding $\Delta_i=V_m\alpha_i$ with the same residual-SVD basis and objective. Table~\ref{tab:shared_per_view} shows that shared steering achieves higher average robust accuracy across the eight fine-grained datasets on CLIP-ViT-B/16 (45.8\% vs.\ 44.9\%),
while learning a single $m$-dimensional coefficient vector instead of one per view.
This result supports the shared correction as an effective cross-view regularizer.

\paragraph{Discriminative Utility of the Residual Subspace.}

We assess whether the residual subspace contains directions useful for classification. Let $g_y$ be the gradient of the reliability-weighted true-class margin $\sum_i w_i[z_{i,y}(\Delta)-\max_{c\neq y}z_{i,c}(\Delta)]$ with respect to the shared shift at $\Delta=0$, where $z_{i,c}$ is the class-$c$ logit of view $i$. 
Its subspace coverage, $C_y(V_m)=\|V_m^\top g_y\|_2^2/\|g_y\|_2^2$, measures the fraction of squared gradient norm captured by the basis.

Table~\ref{tab:steering_analysis} shows higher coverage than rank-matched random bases. Since coverage alone does not establish the effectiveness of entropy minimization, we also compare predictions before and after steering. Steering improves robust accuracy by $3.9$ points on ResNet-50 and $7.4$ points on ViT-B/16, while increasing the average cosine similarity between the original-view feature and its paired clean feature. These results support discriminative steering, without establishing exact clean-feature recovery. Labels and clean features are used only for these post-hoc diagnostics.

\section{Conclusion}

This paper addressed adversarial test-time defense for zero-shot vision-language models by proposing \textbf{Test-time Visual Subspace Steering (T-VSS)}, a lightweight feature-space adaptation method that adjusts attacked visual representations at test time. The central idea is to estimate a compact sample-specific visual subspace from multi-view anchor residuals and to learn a shared, reliability-aware correction inside that subspace. By constraining adaptation to this low-rank geometry, T-VSS turns test-time entropy minimization into structured feature-space steering rather than prompt-space adjustment or dense input-space search.
Extensive experiments across multiple CLIP backbones and diverse benchmarks show that T-VSS improves adversarial robustness under diverse defense-oblivious attacks while preserving competitive clean accuracy.
Further analyses support the discriminative utility of the sample-specific residual subspace and the benefit of shared steering, while demonstrating a favorable robustness--efficiency trade-off against prior test-time defenses.


\section*{LLM Usage Statement}
We used generative AI tools to assist with language editing and translation, code cleanup and refactoring, and the implementation of the proposed method.
We did not use generative AI to develop the core research ideas, design the methodology or experiments, or interpret the experimental results.
All AI-assisted code and text were reviewed by the authors.
The authors take full responsibility for the final manuscript, code, claims,
and other artifacts produced in this work.

{
    \small
    \bibliographystyle{ieeenat_fullname}
    \bibliography{main}
}

\newpage
\clearpage
\setcounter{page}{1}
\appendix
\maketitlesupplementary
\section*{\centering\LARGE Appendix}
\label{sec:appendix}

\startcontents[appendixtoc]
\printcontents[appendixtoc]{l}{1}{\setcounter{tocdepth}{2}}
\addtocontents{toc}{\protect\setcounter{tocdepth}{2}}

\section{More Experimental Details}
\label{app:experimental_details}

\subsection{Datasets}
Table~\ref{tab:datasets} summarizes the number of classes and test samples for all datasets used in our experiments.

\begin{table}[h]
\centering
\caption{Dataset statistics used in the experiments.}
\label{tab:datasets}
\resizebox{.7\linewidth}{!}{%

\begin{tabular}{lcc}
\toprule
Dataset & \# Classes & \# Test \\
\midrule
Caltech101 & 100 & 2,465 \\
Pets & 37 & 3,669 \\
Cars & 196 & 8,041 \\
Flower102 & 102 & 2,463 \\
Aircraft & 100 & 3,333 \\
DTD & 47 & 1,692 \\
EuroSAT & 10 & 8,100 \\
UCF101 & 101 & 3,783 \\
\midrule
ImageNet & 1,000 & 50,000 \\
ImageNet-A & 200 & 7,500 \\
ImageNet-V2 & 1,000 & 10,000 \\
ImageNet-R & 200 & 30,000 \\
ImageNet-S & 1,000 & 50,889 \\
\bottomrule
\end{tabular}
}
\end{table}

\subsection{Augmentation Configuration}

We use the same augmentation pipeline and hyperparameters as R-TPT~\cite{sheng2025r}. For each test sample, we retain the original image as the anchor view and independently generate 63 augmented views. Each augmented view is produced using \texttt{RandomResizedCrop} and \texttt{RandomHorizontalFlip}, followed by AugMix~\cite{hendrycks2020augmix}. The AugMix operation set contains nine transformations: \texttt{AutoContrast}, \texttt{Equalize}, \texttt{Posterize}, \texttt{Rotate}, \texttt{Solarize}, \texttt{ShearX}, \texttt{ShearY}, \texttt{TranslateX}, and \texttt{TranslateY}. We use severity 1 and a mixture width of three augmentation chains. The depth of each chain is sampled independently and uniformly from $\{1,2,3\}$, and the Dirichlet and Beta distributions use concentration parameter $\alpha=1.0$. Together with the unmodified anchor image, this produces the 64-view input set used in our default evaluation.

\subsection{Source of Baseline Results}

Unless otherwise noted, many of the baseline results reported in our tables are taken directly from the original R-TPT~\cite{sheng2025r} and TTP~\cite{ttp} papers when the evaluation setting matches ours in backbone, dataset, and attack protocol. We reformat these numbers only for presentation consistency across tables. When an exact result is not available in the original paper, we reproduce the baseline using the official implementation; such entries are marked with $\dagger$ in the tables.

\begin{table*}[t]
\centering
\caption{Clean (Acc.) and adversarial (Rob.) top-1 accuracy (\%) on eight fine-grained datasets with a TeCoA-pretrained CLIP-ViT-B/32 backbone ($\epsilon=4/255$). Best clean and robust results are highlighted in \textbf{bold} and \textbf{\textcolor{blush}{bold}}, respectively.}
\label{tab:tecoa_pretrained}
\resizebox{\linewidth}{!}{%
\begin{tabular}{lcacacacacacacacaca}
\toprule
\multicolumn{1}{l}{Method}
& \multicolumn{2}{c}{Caltech101}
& \multicolumn{2}{c}{Pets}
& \multicolumn{2}{c}{Cars}
& \multicolumn{2}{c}{Flower102}
& \multicolumn{2}{c}{Aircraft}
& \multicolumn{2}{c}{DTD}
& \multicolumn{2}{c}{EuroSAT}
& \multicolumn{2}{c}{UCF101}
& \multicolumn{2}{c}{Avg.} \\
\cmidrule(lr){2-3}\cmidrule(lr){4-5}\cmidrule(lr){6-7}\cmidrule(lr){8-9}\cmidrule(lr){10-11}\cmidrule(lr){12-13}\cmidrule(lr){14-15}\cmidrule(lr){16-17}\cmidrule(lr){18-19}
\multicolumn{1}{c}{}
& \multicolumn{1}{c}{Acc.} & \multicolumn{1}{c}{Rob.}
& \multicolumn{1}{c}{Acc.} & \multicolumn{1}{c}{Rob.}
& \multicolumn{1}{c}{Acc.} & \multicolumn{1}{c}{Rob.}
& \multicolumn{1}{c}{Acc.} & \multicolumn{1}{c}{Rob.}
& \multicolumn{1}{c}{Acc.} & \multicolumn{1}{c}{Rob.}
& \multicolumn{1}{c}{Acc.} & \multicolumn{1}{c}{Rob.}
& \multicolumn{1}{c}{Acc.} & \multicolumn{1}{c}{Rob.}
& \multicolumn{1}{c}{Acc.} & \multicolumn{1}{c}{Rob.}
& \multicolumn{1}{c}{Acc.} & \multicolumn{1}{c}{Rob.} \\
\midrule
\multicolumn{19}{c}{\textbf{TeCoA-CLIP-ViT-B/32} ($\epsilon=4/255$)} \\
\midrule
CLIP-TeCoA & 79.3 & 44.3 & \bestclean{66.9} & 15.8 & 10.2 & 1.0 & \bestclean{30.8} & 9.0 & 6.6 & 0.5 & 24.5 & 10.7 & \bestclean{14.5} & 10.8 & 34.6 & 6.7 & 33.4 & 12.3 \\
Ensemble & 72.7 & 55.1 & 59.9 & 38.9 & 5.6 & 2.7 & 26.6 & 16.0 & 4.2 & 2.0 & 23.5 & 16.2 & 12.5 & 11.0 & 26.4 & 14.0 & 28.9 & 19.5 \\ 
ET3 & 79.1 & 50.6 & \bestclean{66.9} & 27.2 & 10.3 & 2.0 & 28.9 & 12.1 & 5.3 & 0.9 & 24.0 & 13.2 & 13.4 & 11.2 & \bestclean{37.4} & 16.0 & 33.2 & 16.5 \\
TPT & 79.3 & 52.7 & 65.2 & 27.4 & 9.6 & 2.0 & 27.9 & 12.3 & 6.7 & 1.7 & 25.5 & 14.6 & 12.2 & 11.2 & 34.9 & 10.2 & 32.7 & 16.5 \\
C-TPT & \bestclean{79.8} & 47.3 & 66.1 & 19.5 & \bestclean{10.6} & 1.3 & 29.4 & 10.7 & 6.4 & 0.7 & \bestclean{26.2} & 12.4 & 13.0 & 11.1 & {36.4} & 8.1 & \bestclean{33.5} & 13.9 \\
MTA & 79.7 & 55.7 & 66.2 & 31.2 & 9.0 & 2.5 & 29.1 & 14.0 & 6.5 & 1.6 & 24.4 & 13.5 & 13.3 & 11.2 & 34.6 & 12.5 & 32.9 & 17.8 \\
R-TPT & 76.1 & 60.5 & 63.2 & 40.1 & 7.7 & 3.5 & 26.6 & \bestrob{16.5} & 6.1 & 2.7 & 25.2 & 17.7 & 11.5 & 11.3 & 31.1 & 17.4 & 30.9 & 21.2 \\
T-VSS (Ours) & 77.0 & \bestrob{62.4} & 61.0 & \bestrob{41.3} & 8.7 & \bestrob{4.1} & 25.7 & 16.4 & \bestclean{7.3} & \bestrob{2.9} & 24.5 & \bestrob{18.1} & 11.8 & \bestrob{11.4} & 31.5 & \bestrob{19.6} & 30.9 & \bestrob{22.0} \\
\bottomrule
\end{tabular}%
}
\end{table*}

\begin{table*}[h]
\centering
\caption{Comparison of training-time and test-time defenses on fine-grained classification datasets with pretrained CLIP-ViT-B/32 ($\epsilon=4/255$). Best clean (Acc.) and adversarial (Rob.) results are highlighted in \textbf{bold} and \textbf{\textcolor{blush}{bold}}, respectively. $\dagger$ indicates reproduced results.}

\resizebox{\linewidth}{!}{%
\begin{tabular}{lcacacacacacacacaca}
\toprule
\multicolumn{1}{l}{Method} 
& \multicolumn{2}{c}{Caltech101} 
& \multicolumn{2}{c}{Pets} 
& \multicolumn{2}{c}{Cars} 
& \multicolumn{2}{c}{Flower102} 
& \multicolumn{2}{c}{Aircraft} 
& \multicolumn{2}{c}{DTD} 
& \multicolumn{2}{c}{EuroSAT} 
& \multicolumn{2}{c}{UCF101} 
& \multicolumn{2}{c}{Avg.} \\
\cmidrule(lr){2-3}\cmidrule(lr){4-5}\cmidrule(lr){6-7}\cmidrule(lr){8-9}\cmidrule(lr){10-11}\cmidrule(lr){12-13}\cmidrule(lr){14-15}\cmidrule(lr){16-17}\cmidrule(lr){18-19}
\multicolumn{1}{c}{} 
& \multicolumn{1}{c}{Acc.} & \multicolumn{1}{c}{Rob.} 
& \multicolumn{1}{c}{Acc.} & \multicolumn{1}{c}{Rob.} 
& \multicolumn{1}{c}{Acc.} & \multicolumn{1}{c}{Rob.} 
& \multicolumn{1}{c}{Acc.} & \multicolumn{1}{c}{Rob.} 
& \multicolumn{1}{c}{Acc.} & \multicolumn{1}{c}{Rob.} 
& \multicolumn{1}{c}{Acc.} & \multicolumn{1}{c}{Rob.} 
& \multicolumn{1}{c}{Acc.} & \multicolumn{1}{c}{Rob.} 
& \multicolumn{1}{c}{Acc.} & \multicolumn{1}{c}{Rob.} 
& \multicolumn{1}{c}{Acc.} & \multicolumn{1}{c}{Rob.} \\
\midrule
\multicolumn{19}{c}{\textbf{CLIP-ViT-B/32} ($\epsilon=4/255$)} \\
\midrule
CLIP & 91.4 & 0.2 & 85.1 & 0.0 & 60.1 & 0.0 & 64.0 & 0.0 & 18.1 & 0.0 & 43.0 & 0.0 & 35.8 & 0.0 & 61.6 & 0.0 & 57.4 & 0.0 \\

\multicolumn{17}{l}{\textbf{Training-time Defense Methods}} \\

TeCoA & 79.3 & 78.0 & 66.9 & 63.7 & 10.2 & 9.1 & 30.8 & 28.9 & 6.6 & 5.9 & 24.5 & 24.0 & 14.5 & 14.3 & 34.6 & 33.4 & 33.4 & 32.2 \\
FARE & 86.3 & \textcolor{blush}{\textbf{85.4}} & 76.7 & \textcolor{blush}{\textbf{73.8}} & 39.2 & 34.4 & 37.0 & 34.0 & 9.5 & 8.5 & 28.3 & 27.3 & 16.6 & 16.3 & 44.2 & 41.9 & 42.2 & 40.2 \\

APT & 10.7 & 0.4 & 10.0 & 0.2 & 1.5 & 0.1 & 0.9 & 0.2 & 2.6 & 0.5 & 9.0 & 0.1 & 7.8 & 6.7 & 3.7 & 0.2 & 5.8 & 1.0 \\
APT+TeCoA & 81.4 & 80.2 & 66.7 & 63.9 & 20.8 & 18.9 & 42.5 & 40.4 & 5.2 & 5.0 & 35.2 & \textcolor{blush}{\textbf{33.7}} & 29.3 & \textcolor{blush}{\textbf{29.2}} & 40.2 & 39.4 & 40.2 & 38.8 \\
\multicolumn{17}{l}{\textbf{Test-time Defense Methods}} \\
TTC & 86.5 & 22.7 & 83.5 & 11.8 & 48.1 & 2.3 & 64.3 & 3.2 & 18.2 & 1.0 & 37.3 & 4.7 & \textbf{53.0} & 3.0 & 62.6 & 6.1 & 56.7 & 6.9 \\

ET3\dag & 88.6 & 26.9 & 82.0 & 9.6 & 53.2 & 0.5 & 59.0 & 2.8 & 14.7 & 0.2 & 35.8 & 6.4 & 23.4 & 1.5 & 57.2 & 4.6 & 51.8 & 6.6 \\

Ensemble & 88.2 & 74.9 & 75.0 & 52.5 & 51.7 & 25.9 & 58.1 & 36.1 & 16.4 & 7.9 & 39.8 & 28.6 & 30.8 & 11.9 & 54.9 & 36.9 & 51.9 & 34.3 \\
MTA & \textbf{92.0} & 76.3 & \textbf{86.3} & 53.6 & \textbf{63.4} & 26.4 & \textbf{64.4} & 36.5 & \textbf{20.2} & 8.2 & \textbf{43.8} & 28.8 & 34.6 & 11.3 & \textbf{63.3} & 39.1 & \textbf{58.5} & 35.0 \\
R-TPT & 90.6 & 76.4 & 84.5 & 55.8 & 63.1 & 28.4 & 62.6 & 37.6 & 19.1 & 9.2 & 42.1 & 29.1 & 32.0 & 5.1 & 62.8 & 41.0 & 57.1 & 35.3 \\
TTP & 90.9 & 81.8 & 84.7 & 61.0 & 59.8 & 29.8 & 63.6 & 42.0 & 18.0 & 10.3 & 42.8 & 32.2 & 35.6 & 14.1 & 61.3 & \textcolor{blush}{\textbf{46.6}} & 57.1 & 39.7 \\
T-VSS (Ours) & 91.7 & 76.6 & 85.3 & 63.2 & 61.2 & \textcolor{blush}{\textbf{43.8}} & 62.4 & \textcolor{blush}{\textbf{45.2}} & 19.6 & \textcolor{blush}{\textbf{15.8}} & 43.3 & 33.1 & 29.8 & 6.6 & 61.7 & 43.8 & 56.9 & \textcolor{blush}{\textbf{41.0}} \\
\bottomrule
\end{tabular}%
}
\label{tab:vitb32}
\end{table*}

\subsection{Diagnostic Metric Definitions}
\label{app:diagnostic_metrics}

For Table~\ref{tab:paired_feature_analysis}, let $M=N+1$ be the number of views and $\delta_i=f_i^{\mathrm{adv}}-f_i^{\mathrm{clean}}$. Cross-view alignment is
\begin{equation}
A_{\mathrm{shift}}=\frac{1}{M(M-1)}\sum_{i\neq j}\cos(\delta_i,\delta_j).
\end{equation}
With $\bar{\delta}=M^{-1}\sum_i\delta_i$, the shared-energy ratio is
\begin{equation}
E_{\mathrm{shared}}=
\frac{\|\bar{\delta}\|_2^2}
{M^{-1}\sum_i\|\delta_i\|_2^2}.
\end{equation}
Clean and adversarial ranks are the minimum numbers of singular directions required to explain $90\%$ of the corresponding anchor-relative residual energy.

\section{Additional Experiments}

\subsection{Results under Robust Pretrained Backbone}

Table~\ref{tab:tecoa_pretrained} evaluates whether T-VSS remains effective when the underlying CLIP-ViT-B/32 encoder is already robust-pretrained with TeCoA~\cite{mao2023understanding}. The answer is affirmative: T-VSS achieves the best average robust accuracy at $22.0\%$, improving over the robust-pretrained baseline itself by $9.7$ points and over the strongest test-time baseline, R-TPT, by $0.8$ points. The gain is also consistent across individual datasets, where T-VSS attains the best robust accuracy on seven of the eight benchmarks. These results suggest that the proposed feature-space correction is complementary to training-time robustness and can further improve an already strengthened visual encoder without any additional fine-tuning or retraining.

\begin{table*}[ht]
\centering
\caption{Adversarial accuracy (\%) under additional attacks on Flower102 and DTD with CLIP-ViT-B/16. Best results are highlighted in \textbf{\textcolor{blush}{bold}}. }
\label{tab:additional_attacks_vitb16}
\resizebox{.8\linewidth}{!}{%
\begin{tabular}{laaaaaa}
\toprule
\multirow{2}{*}{Method} & \multicolumn{3}{c}{Flower102} & \multicolumn{3}{c}{DTD} \\
\cmidrule(lr){2-4}\cmidrule(lr){5-7}
& \multicolumn{1}{c}{AutoAttack} &  \multicolumn{1}{c}{APGD-CE} & \multicolumn{1}{c}{APGD-DLR} & \multicolumn{1}{c}{AutoAttack} & \multicolumn{1}{c}{APGD-CE} & \multicolumn{1}{c}{APGD-DLR} \\
\midrule
CLIP & 0.0 &  0.0 & 0.0 & 0.0 &  0.0 & 0.0 \\
R-TPT & 39.2 &  39.5 & 46.5 & 32.4 &  32.4 & 34.6 \\
TTP & 26.7 &  38.9 & 27.1 & 22.3 &  26.0 & 22.6 \\
T-VSS (Ours) & \bestrob{45.1} &  \bestrob{45.1} & \bestrob{51.6} & \bestrob{33.3} & \bestrob{45.1} & \bestrob{36.2} \\
\bottomrule
\end{tabular}%
}
\end{table*}

\subsection{Additional CLIP Backbone Results}
Table~\ref{tab:vitb32} further evaluates T-VSS on CLIP-ViT-B/32 and broadens the comparison to both training-time and test-time defenses. The overall trend remains consistent: T-VSS achieves the best average robust accuracy at $41.0\%$, outperforming the strongest test-time baseline, TTP, by $1.3$ points and the strongest training-time baseline, FARE~\cite{schlarmann2024robust}, by $0.8$ points. The gains are particularly clear on Cars, Flower102, and Aircraft, with an improvement in the overall average, showing that the proposed feature-space correction transfers effectively to this additional backbone. Although some training-time defenses remain competitive on individual datasets or in clean accuracy, they require robust pretraining or adversarial fine-tuning. By contrast, T-VSS delivers the strongest overall robustness without any additional training, reinforcing the practical advantage of direct test-time feature correction.

\begin{table*}[t]
\begin{minipage}[t]{0.48\textwidth}
\vspace{0pt}
\centering
\includegraphics[width=0.75\linewidth]{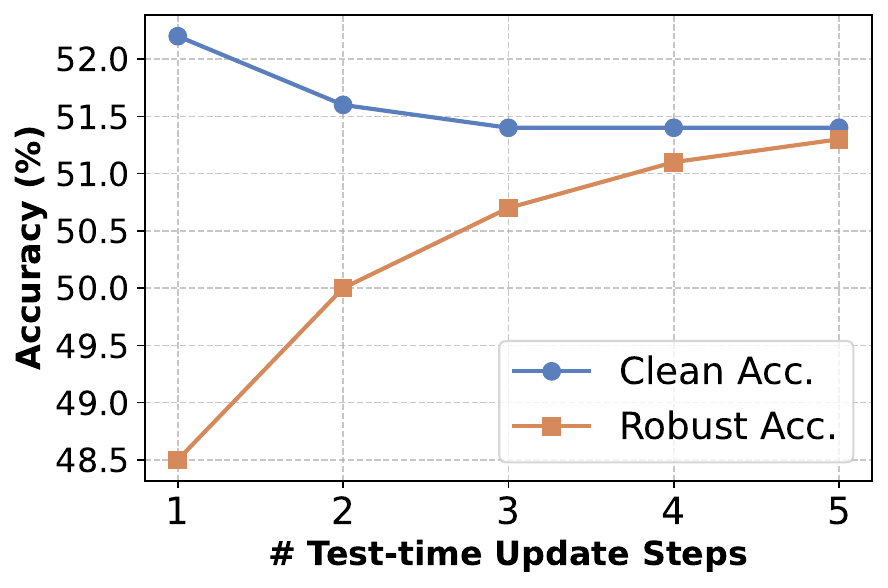}
\captionof{figure}{Effect of the number of test-time update steps on average clean (Acc.) and adversarial (Rob.) accuracy (\%) over the eight fine-grained datasets with CLIP-ResNet-50.}
\label{fig:update_step_ablation}
\end{minipage}
\hfill
\begin{minipage}[t]{0.48\textwidth}
\vspace{0pt}
\centering
\captionof{table}{Mean $\pm$ standard deviation of clean and adversarial (robust) accuracy (\%) averaged over the eight fine-grained datasets across three independent runs with different random seeds. T-VSS shows consistently low variance across all backbones.}
\label{tab:performance_mean_std}
\begin{tabular}{lca}
\toprule
Backbone & Acc.& \multicolumn{1}{c}{Rob.} \\
\midrule
ResNet-50 & $52.2_{\pm 0.1}$ & $48.5_{\pm 0.1}$ \\
ViT-B/16  & $60.4_{\pm 0.1}$ & $45.8_{\pm 0.2}$ \\
ViT-B/32  & $56.9_{\pm 0.1}$ & $41.0_{\pm 0.1}$ \\
ViT-L/14  & $68.1_{\pm 0.1}$ & $53.8_{\pm 0.1}$ \\
\bottomrule
\end{tabular}
\end{minipage}
\end{table*}

\subsection{Evaluation under Additional Attacks}

\label{appendix:additional_attack}

Table~\ref{tab:additional_attacks_vitb16} extends the evaluation beyond the PGD setting used in the main paper by considering stronger composite and first-order attacks on CLIP-ViT-B/16. Across both Flower102 and DTD, T-VSS remains consistently robust under AutoAttack~\cite{autoattack}, APGD-CE, and APGD-DLR, achieving the best adversarial accuracy in every reported setting. The gains are especially clear on Flower102, where T-VSS improves over the strongest prior baseline by $5.9$ points under AutoAttack, $5.6$ points under APGD-CE, and $5.1$ points under APGD-DLR.
On DTD, the advantage is smaller but still consistent under AutoAttack and APGD-DLR, while under APGD-CE
%
%
the advantage is substantially larger: T-VSS increases robust accuracy from $32.4\%$ for R-TPT to $45.1\%$.
These results strengthen the main claim of the paper: the benefit of T-VSS does not depend narrowly on a single PGD configuration, but persists under diverse optimization-based attacks, suggesting that the proposed low-rank feature correction provides a more stable adaptation mechanism than prior prompt-space or input-space defenses.

\subsection{Effect of the Number of Update Steps}
Figure~\ref{fig:update_step_ablation} shows the effect of increasing the number of test-time update steps on the average clean and adversarial accuracy over the eight fine-grained datasets with CLIP-ResNet-50. Adversarial accuracy improves steadily as more update steps are used, increasing from $48.5\%$ with one step to $51.3\%$ with five steps. This gain comes with a small clean-accuracy cost: clean accuracy drops from $52.2\%$ to $51.4\%$ and then largely saturates after three steps. The overall trend highlights a clear clean--robustness trade-off, where additional optimization steps can further strengthen low-rank feature-space adaptation under attack, while the default one-step setting remains attractive when inference efficiency and clean performance are both important.

\subsection{Stability Across Random Seeds}

Table~\ref{tab:performance_mean_std} reports the mean and standard deviation of clean and adversarial accuracy averaged over the eight fine-grained datasets across three independent runs with different random seeds. T-VSS exhibits uniformly low variance across all backbones, with fluctuations of at most $0.2$ points in both clean and adversarial accuracy. This stability indicates that the method is not sensitive to seed-specific initialization or stochastic view generation at test time. In other words, the gains reported in the main paper are not driven by a favorable run, but are reproduced consistently across independent trials.

\subsection{Comparison with DBD}

\begin{table*}[t]
\centering
\caption{Average selected rank $m$ across eight fine-grained datasets under the default 64-view protocol (one original image + 63 augmented views). Adversarial rows use the same backbone-specific PGD attacks as in the main paper. Since T-VSS optimizes only the $m$-dimensional coefficient vector, the selected rank is also the number of sample-wise learnable parameters.}
\label{tab:selected_rank}
\resizebox{\linewidth}{!}{%
\begin{tabular}{llccccccccc}
\toprule
Setting & Model & Caltech101 & Pets & Cars & Flower102 & Aircraft & DTD & EuroSAT & UCF101 & Avg. \\
\midrule
\multirow{4}{*}{\textbf{Clean}}
& ResNet-50 & 13.3 & 14.9 & 14.2 & 14.1 & 16.6 & 12.1 & 8.3 & 13.7 & 13.4 \\
& ViT-B/16  & 12.3 & 13.8 & 13.4 & 12.6 & 17.0 & 12.6 & 7.9 & 12.6 & 12.8 \\
& ViT-B/32  & 14.0 & 15.5 & 14.8 & 14.6 & 17.7 & 13.5 & 9.8 & 14.4 & 14.3 \\
& ViT-L/14  & 14.8 & 15.5 & 14.5 & 14.6 & 16.5 & 15.4 & 11.7 & 14.2 & 14.6 \\
\midrule
\multirow{4}{*}{\textbf{Adversarial}}
& ResNet-50 & 6.8 & 6.0 & 5.9 & 5.4 & 6.3 & 5.4 & 3.6 & 6.5 & 5.7 \\
& ViT-B/16  & 2.0 & 1.5 & 1.9 & 1.6 & 2.2 & 1.5 & 1.4 & 2.2 & 1.8 \\
& ViT-B/32  & 3.1 & 2.6 & 3.0 & 3.2 & 3.3 & 2.5 & 2.0 & 3.6 & 2.9 \\
& ViT-L/14  & 2.2 & 1.9 & 2.1 & 1.7 & 2.9 & 2.1 & 2.3 & 2.4 & 2.2 \\
\bottomrule
\end{tabular}%
}
\end{table*}

\begin{table}[t]
\centering
\caption{Case-study comparison with DBD on Caltech101 under standard and robustly pretrained ViT-B/32 backbones. DBD is strongest on standard CLIP-ViT-B/32, whereas T-VSS attains better clean accuracy and slightly higher adversarial accuracy on TeCoA-CLIP-ViT-B/32.
}
\label{tab:dbd_comparison}
\resizebox{.6\linewidth}{!}{%
\begin{tabular}{lca}
\toprule
Method & Acc. & \multicolumn{1}{c}{Rob.} \\
\midrule
\multicolumn{3}{c}{\textbf{CLIP-ViT-B/32} ($\epsilon=4/255$)} \\
\midrule
CLIP & 91.4 & 0.2 \\
R-TPT & 90.6 & 76.4 \\
DBD & 90.4 & \bestrob{98.8} \\
T-VSS (Ours) & \bestclean{91.7} & 76.6 \\
\midrule
\multicolumn{3}{c}{\textbf{TeCoA-CLIP-ViT-B/32} ($\epsilon=4/255$)} \\
\midrule
CLIP-TeCoA & \bestclean{79.3} & 44.3 \\
R-TPT & 76.1 & 60.5 \\
DBD & 70.9 & 60.9 \\
T-VSS (Ours) & 77.0 & \bestrob{62.4} \\
\bottomrule
\end{tabular}
}
\end{table}

DBD~\cite{liu2026adversarial} is a strong concurrent training-free defense that also reconstructs visual features at test time, but it follows a different inference regime from the optimization-based methods in our main benchmark. Specifically, DBD estimates a single defense direction from transformed views and applies a DB-score-based thresholded reconstruction rule with validation-calibrated hyperparameters, whereas T-VSS performs sample-wise test-time optimization in a low-rank visual subspace without thresholded routing. Table~\ref{tab:dbd_comparison} shows a case study on Caltech101. On standard CLIP-ViT-B/32, DBD is substantially stronger than all optimization-based baselines, which is consistent with the effectiveness of its calibrated single-direction reconstruction on the standard backbone. For the TeCoA-CLIP-ViT-B/32 result, we apply DBD using the DB-score threshold reported in the original paper, without additional recalibration for the robust backbone. Under this setting, DBD still improves adversarial accuracy over the TeCoA backbone, but its advantage becomes much smaller and it achieves lower clean accuracy and slightly lower adversarial accuracy than T-VSS. We do not claim that DBD cannot be improved further with backbone-specific retuning; rather, this case study suggests that its fixed thresholded rule may transfer less directly across backbones when robust pretraining changes the underlying feature geometry. By contrast, T-VSS uses sample-specific low-rank optimization without hard thresholding, which may help it transfer more favorably to the robustly pretrained backbone considered here.

\subsection{Selected Rank $m$ and Number of Learnable Parameters}

Table~\ref{tab:selected_rank} reports the average selected rank $m$ across the eight fine-grained datasets under the default 64-view protocol, which consists of one original image and 63 augmented views. For adversarial evaluation, the rank is measured under the same backbone-specific PGD attacks used in the main paper. Since T-VSS optimizes only the $m$-dimensional coefficient vector $\alpha$, the selected rank is exactly the number of sample-wise learnable parameters at test time. Even though T-VSS uses 63 augmented views, the average rank on adversarial examples remains very small: $5.7$ for ResNet-50, $1.8$ for ViT-B/16, $2.9$ for ViT-B/32, and $2.2$ for ViT-L/14. In other words, T-VSS typically performs sample-wise adaptation with only a handful of learnable coefficients, far fewer than prior optimization-based defenses such as R-TPT and TTP, which optimize higher-dimensional prompt or input variables at test time.
The lower selected ranks under attack indicate stronger energy concentration in the anchor-relative residuals, allowing T-VSS to parameterize its shared correction with only a few coefficients.

\section{Limitations and Future Work}
\label{appendix:eot}

Like other augmentation-driven test-time defenses considered in this work, T-VSS relies on stochastic multi-view augmentation at inference time. While this mechanism is effective under the defense-oblivious threat model used in our main experiments, the augmentation and adaptation pipeline can itself become an attack surface when the adversary explicitly accounts for the defense. We assess this limitation using defense-aware PGD with Expectation over Transformation (EOT)~\cite{eot}, which differentiates through the stochastic augmentation distribution and the test-time optimization procedure. We evaluate R-TPT, TTP, and T-VSS on Flower102 and DTD using both CLIP-ResNet-50 and CLIP-ViT-B/16.
For both backbones, the EOT attack and defended inference use a matched 64-view budget. At each PGD step, the gradient is estimated from one stochastic defended forward pass comprising the original image and 63 augmented views.

\begin{table}[t]
\centering
\caption{Robust accuracy (\%) under defense-aware EOT-PGD. Best results for each backbone are in \textbf{bold}.}
\label{tab:eot_pgd}
\resizebox{.8\linewidth}{!}{%
\begin{tabular}{llaa}
\toprule
Backbone & Method & \multicolumn{1}{c}{Flower102} & \multicolumn{1}{c}{DTD} \\
\midrule
\multirow{3}{*}{ResNet-50}
& R-TPT & 25.8 & 23.1 \\
& TTP & 1.5 & 4.4 \\
& T-VSS (Ours) & \bestrob{40.1} & \bestrob{26.7} \\
\midrule
\multirow{3}{*}{ViT-B/16}
& R-TPT & 0.0 & 2.1 \\
& TTP & 0.9 & 1.3 \\
& T-VSS (Ours) & \bestrob{4.0} & \bestrob{7.2} \\
\bottomrule
\end{tabular}%
}
\end{table}

As shown in Table~\ref{tab:eot_pgd}, T-VSS attains the highest robust accuracy in all four backbone--dataset settings. Its advantage is substantial on ResNet-50 and remains consistent on ViT-B/16. Nevertheless, absolute robustness on ViT-B/16 remains below 8\% for all methods, confirming that differentiating through multi-view adaptation poses a serious challenge under the stronger attack. Improving robustness to adaptive attacks while retaining the benefits of stochastic multi-view inference is therefore an important direction for future work.

\section{Ethics Statement}
This work aims to improve the reliability of zero-shot vision-language models under adversarial perturbations. Stronger test-time defense can be beneficial in high-stakes settings such as medical decision support and autonomous perception, where small input corruptions may otherwise cause harmful errors. However, improved robustness is not a guarantee of safety and should not be over-interpreted, especially because defense methods may still fail under stronger adaptive attacks. In addition, robustness research is inherently dual-use, since it may also inform the design of stronger attacks. We therefore view T-VSS as a complementary safety mechanism that should be paired with rigorous evaluation and additional safeguards in real-world deployment.

\end{document}